\documentclass[sigconf, nonacm]{acmart}

\usepackage{float}
\usepackage{booktabs}
\usepackage{amsmath}
\usepackage{amsthm}
\usepackage{graphicx}
\usepackage{balance} 
\usepackage{needspace} 
\usepackage{booktabs}
\usepackage{algorithm}
\usepackage{algpseudocode}
\usepackage{diagbox}
\usepackage{tabularx}
\usepackage{verbatim}
\usepackage{balance}

\usepackage[colorinlistoftodos]{todonotes}
\newtheorem{definition}{Definition}

\begin{document}
\title{Raising the ClaSS of Streaming Time Series Segmentation}


\author{Arik Ermshaus}
\affiliation{%
  \institution{Humboldt-Universit\"at zu Berlin}
  \city{Berlin}
  \country{Germany}
}
\email{ermshaua@informatik.hu-berlin.de}

\author{Patrick Sch\"afer}
\affiliation{%
  \institution{Humboldt-Universit\"at zu Berlin}
  \city{Berlin}
  \country{Germany}
}
\email{patrick.schaefer@hu-berlin.de}

\author{Ulf Leser}
\affiliation{%
  \institution{Humboldt-Universit\"at zu Berlin}
  \city{Berlin}
  \country{Germany}
}
\email{leser@informatik.hu-berlin.de}

\begin{abstract}
  Ubiquitous sensors today emit high frequency streams of numerical measurements that reflect properties of human, animal, industrial, commercial, and natural processes. Shifts in such processes, e.g. caused by external events or internal state changes, manifest as changes in the recorded signals. The task of streaming time series segmentation (STSS) is to partition the stream into consecutive variable-sized segments that correspond to states of the observed processes or entities. The partition operation itself must in performance be able to cope with the input frequency of the signals. We introduce ClaSS, a novel, efficient, and highly accurate algorithm for STSS. ClaSS assesses the homogeneity of potential partitions using self-supervised time series classification and applies statistical tests to detect significant change points (CPs). In our experimental evaluation using two large benchmarks and six real-world data archives, we found ClaSS to be significantly more precise than eight state-of-the-art competitors. Its space and time complexity is independent of segment sizes and linear only in the sliding window size. We also provide ClaSS as a window operator with an average throughput of $1k$ data points per second for the Apache Flink streaming engine.
\end{abstract}

\settopmatter{printfolios=true}
\maketitle

\sloppy

\section{Introduction}

Over the past two decades, the decreasing costs of sensors and the growing digitalization of industry, science, and society has led to an enormous increase in applications that analyse streams of sensor recordings. For example, modern smartphones contain inertial measurement units (IMUs) with triaxial accelerometers, gyroscopes, and magnetometers that can track human activities~\cite{Baos2015DesignIA}. Seismology relies on globally distributed stations to provide high-resolution waveform recordings used for earthquake detection and early warning~\cite{Woollam2022SeisBenchATF}. In cardiology, electrocardiographs (ECG) capture heart beats from subjects over long periods of time to obtain insights into cardiac dynamics such as arrhythmias~\cite{Moody2001TheIO}. Regardless of the domain, the underlying sensors emit continuous sequences of real-valued measurements at a given frequency, called sensor data (series) or \emph{time series} (TS). The literature offers a rich selection of technologies to store, manage, analyse, visualize and search in collections of TS~\cite{Liakos2022ChimpEL, Adams2020MonarchGP, Echihabi2022HerculesAD, Zhang2022PARROTPC, Wen_Gao_Song_Sun_Xu_Zhu_2019, Yeh2016MatrixPI}. Common basic operations are the detection of unusual stretches called \emph{anomalies}~\cite{Paparrizos2022TSBUADAE}, of repetitive structures called \emph{motifs}~\cite{Schfer2022MotifletsS}, and of homogeneous subsequences called \emph{segments}~\cite{Gharghabi2018DomainAO}.

TS methods can be broadly classified into batch or streaming. Methods for the batch analysis of TS, used in applications such as gait or sleep stage analysis~\cite{Truong2019SelectiveRO,Kemp2000AnalysisOA}, can largely ignore latency, runtime and memory requirements and use complex preprocessing based on global statistics (e.g. frequency filtering or signal decomposition). This is different in the streaming case, where infinite TS must be processed in real-time relative to the measurement frequency and where the complexity of operations must not depend on the length of sequences~\cite{Verma2017ASO}. 

This is especially unfortunate for the task of TS segmentation (TSS), a common preprocessing step between data collection~\cite{Reiss2012CreatingAB} and knowledge discovery from TS~\cite{Matsubara2014AutoPlaitAM}. TSS allows inferring the latent states of an underlying process by analysing sensor measurements, as signal shifts from one segment to another are assumed to be caused by state changes in the process being monitored, such as a transition from one human activity to another or from one machine state to another. In the batch case, TSS aims to partition a given TS into consecutive regions such that each region is homogeneous in itself yet sufficiently different from the neighbouring regions. It is typically performed by focusing on the detection of change points (CPs) separating segments~\cite{Aminikhanghahi2016ASO}. State-of-the-art methods for TSS rely on global statistics of the TS, value distributions, densities or learned features~\cite{Truong2019SelectiveRO}, and often exhibit a high computational complexity. Recent accurate contributions, e.g. FLUSS~\cite{Gharghabi2018DomainAO} or ClaSP~\cite{Ermshaus2022WSS}, are quadratic in runtime regarding the TS length.

For the streaming case such statistics are not available and such complexities clearly are not feasible, i.e., for segmenting streams of TS (STSS)~\cite{Pan1985ARQ, Hwang2010ASO, Munchmeyer2020TheTE}. Real-time processing is essential for STSS, yet challenging. For instance, IoT devices may emit measurements with hundreds of Hertz (Hz)~\cite{Baos2014mHealthDroidAN,Schmidt2018IntroducingWA,Woollam2022SeisBenchATF,Goldberger2000PhysioBankPA}. STSS requires algorithms that process data points faster than they arrive, utilizing only a constant amount of memory. The problem was first formulated by Kifer et al. in the context of change detection~\cite{Kifer2004DetectingCI}. The basic approach is to constrain the analysis to the last $d$ observations using a sliding window and to continuously emit detected CPs, which each defines the end of a segment~\cite{Aminikhanghahi2016ASO}. Following this seminal work, many follow-up methods considered stream segmentation as a part of IoT workflows~\cite{Wan2015DynamicSE, Cho2015AutomaticSD, Dbski2021AdaptiveSO} or studied drift or CP detection~\cite{Gama2014ASO,Aminikhanghahi2016ASO}. The best methods, however, only work on temporal data with a suitable value distribution (e.g. BOCD~\cite{adams2007bayesian}), can only detect very limited change or drift types (e.g. NEWMA~\cite{Keriven2018NEWMAAN}), or rely on thresholding as segmentation procedures (e.g. FLOSS~\cite{Gharghabi2018DomainAO}), which is not robust for real-world signals.

\begin{figure}[t]
    \includegraphics[width=1.0\columnwidth]{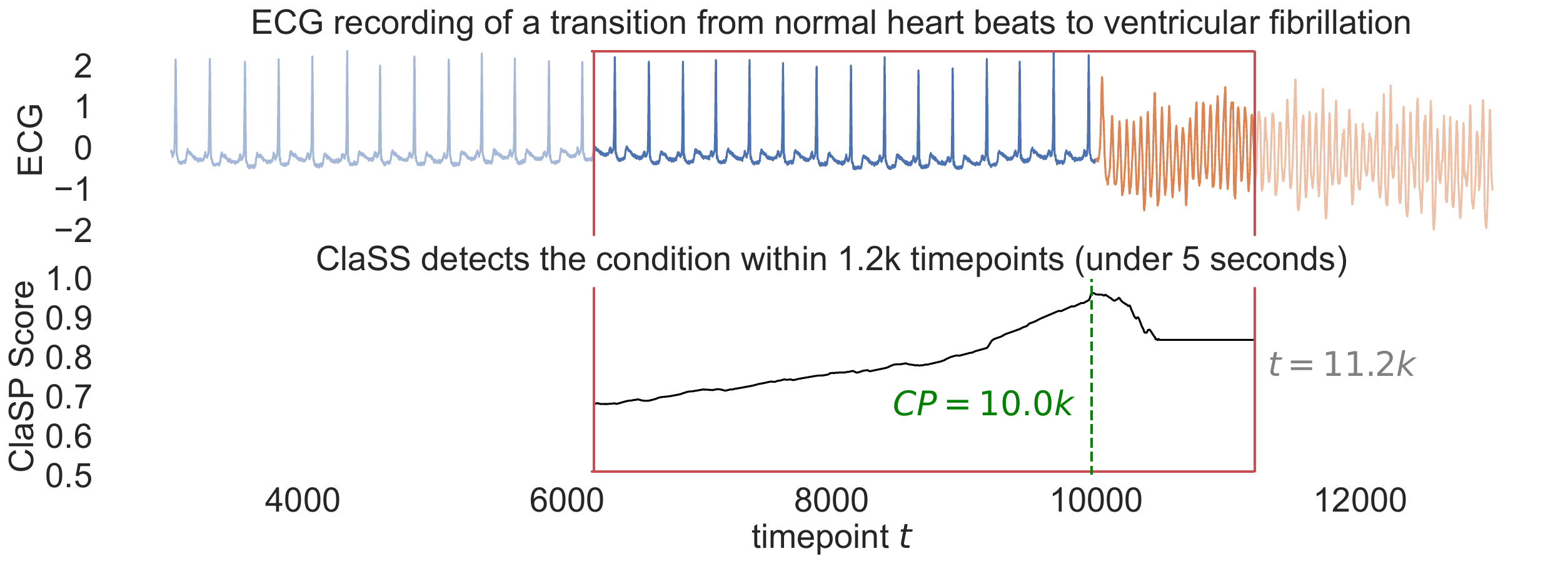}
    \caption{An electrocardiogram (ECG) recording of a human subject demonstrating the transition from normal heartbeats (in blue) to ventricular fibrillations (in orange)~\cite{Nolle1986crei}. The ClaSS algorithm continuously scores the TS stream within a sliding window (shown in red), and at $t = 11.2k$ a significant change in the signal shape is detected and immediately reported to the user. This split effectively divides the stream into a fully processed segment and one that evolves.\label{fig:class_example}
    }
\end{figure}

We present ClaSS (Classification Score Stream), a domain-agnostic, highly accurate and efficient algorithm that approaches STSS as a self-supervised learning problem. It continuously scores the homogeneity of hypothetical sliding window split points and identifies statistically significant CPs using hypothesis testing to find the ends and beginnings of segments. Notably, the algorithm only scores the last detected segment and tests if it should be further split into two, thereby reducing model complexity and saving runtime. The core of this approach lies in an efficient calculation of ClaSP (Classification Score Profile)~\cite{Ermshaus2022ClaSP}, which was originally constrained to batch analysis. ClaSS achieves much higher efficiency than ClaSP, as necessary for the streaming case, by efficiently cross-validating a novel streaming $k$-nearest neighbour ($k$-NN) that re-uses the results of calculations from previous overlapping sliding windows. Time and space complexity of ClaSS are both linearly dependent only on the sliding window size, thus fulfilling the requirements of STSS. While some competitors exhibit sublinear sliding window update runtimes surpassing ClaSS, their segmentation models are restricted to basic methods (e.g. statistical parameter deviation), limiting their accuracy. Conversely, ClaSS trades this runtime gap to incorporate more advanced data mining techniques, resulting in significantly higher accuracy at sufficient speeds.

Figure~\ref{fig:class_example} exemplifies how ClaSS segments a sliding window into homogeneous regions. The data set~\cite{Nolle1986crei} shows an ECG recording sampled at 250 Hz from a human subject, who experienced ventricular fibrillations after 40 seconds ($t=10k$). The global maximum in the profile (Figure~\ref{fig:class_example} bottom) captures the start of the condition, and is detected and reported as a significant change around 5 seconds after the ventricular fibrillations begin ($t=11.2k$); dividing the stream into segments with normal and abnormal cardiac activity. 

Specifically, this paper makes the following contributions:

\begin{enumerate}
    \item We introduce ClaSS, a novel, efficient and domain-agnostic method for STSS, which scores sliding windows using self-supervised TS classification to detect and report statistically significant CPs with low latency. The scoring process annotates the sliding window with the likelihood of CPs. Besides being necessary for STSS, this makes it easy for humans to understand and suitable for decision-support systems. 

    \item We present two technical advancements that enable ClaSS to meet the stringent performance criteria for STSS: the first exact streaming TS $k$-nearest neighbour ($k$-NN) algorithm that runs in $\mathcal{O}(k \cdot d)$ for a single sliding window (of length $d$) update, substantially improving upon the current-best $\mathcal{O}((k + \log d) \cdot d)$ solution~\cite{Gharghabi2018DomainAO}, and a novel algorithm for cross-validating a self-supervised $k$-NN classifier in $\mathcal{O}(d)$, outperforming the prior best $\mathcal{O}(d^2)$ approach~\cite{Ermshaus2022ClaSP}.

    \item We analysed the accuracy of ClaSS using $592$ real-world TS from two benchmarks and six experimental studies. Compared to eight state-of-the-art competitors (FLOSS, DDM, ChangeFinder, NEWMA, BOCD, HDDM, ADWIN and a sliding window baseline), ClaSS significantly outperforms all other competitors, exhibits the highest overall segmentation accuracy and improves the state of the art by $13.7$ pp (percentage points).

\end{enumerate}

We make all of our used source codes, including a stand-alone Python implementation and a comparably fast Apache Flink window operator of ClaSS, the evaluation framework, Jupyter Notebooks, as well as all experiment data and visualizations openly available on our supporting website~\cite{ClaSSWebpage} to foster the reproducibility of our findings and replicability for follow-up works. The rest of the paper is structured as follows. In Section 2, we provide background and definitions for this work. Section 3 introduces our ClaSS algorithm and its components. Section 4 describes results from an extensive experimental evaluation and Section 5 details related work. Finally, we conclude in Section 6.

\section{Background and Definitions}

This section formally introduces the concepts of time series streams, sliding windows, subsequences, the streaming time series segmentation (STSS) problem and self-supervision. We also briefly recapitulate the idea behind ClaSP~\cite{Ermshaus2022ClaSP} upon which ClaSS is based.

\begin{definition}
A TS \emph{stream} $S$ produces a new real-valued data point $t_{\tau} \in \mathbb{R}$ at evenly spaced time intervals, which enqueues in a continuous sequence $<\dots,t_{\tau-1},t_{\tau}>$ of values. The data points are also called observations or measurements.
\end{definition}

The main characteristic of a TS stream is its infinite length. However, in practice, only a finite number of measurements can be stored and processed at any time (i.e., $t_{\tau}$). This necessitates efficient data mining techniques that can quickly analyse incoming data. We focus on univariate streams that are sampled at equidistant time stamps (e.g. 50 Hz), with the same temporal duration between consecutive measurements. It is worth emphasizing that any finitely long TS can be treated as a TS stream and analysed accordingly.

\begin{definition}
Given a TS stream $S$, a \emph{sliding window} $S_{\tau-d+1,\tau}$ is a buffer of size $d$ that stores the latest $d$ data points produced by $S$. As a new data point appears in $S$, the corresponding $S_{\tau-d+1,\tau}$ expels the oldest observation and appends the youngest one.
\end{definition}

\begin{figure}[t]
    \includegraphics[width=1.0\columnwidth]{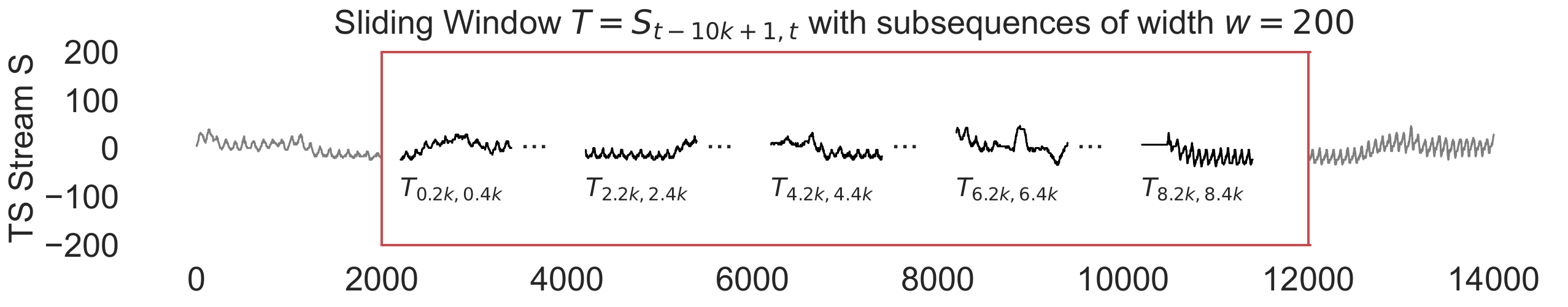}
    \caption{A TS stream $S$ from which the last $d=10k$ observations are buffered in a sliding window $T = S_{\tau-d+1,\tau}$, depicted as the red frame. Older (or yet to arrive) data points are greyed out. The sliding window is further cut into subsequences of width $w = 200$, to be analysed for segmentation.\label{fig:stream_example}
    }
\end{figure}

The size of the window $d$ is a hyper-parameter that will be discussed in Subsection~\ref{sec:hyper-param}. This choice directly affects the amount of information and thus the runtime and memory of the algorithms applied to $S_{\tau-d+1,\tau}$. For an example of a sliding window, see Figure~\ref{fig:stream_example}. Note that we represent a window as a finite TS $T = S_{\tau-d+1,\tau}$ of size $d$, from which we can access values at offsets $[1 \dots d]$.  

\begin{definition}
Given a sliding window $T = S_{\tau-d+1,\tau}$, a \emph{subsequence} $T_{s,e}$ of $T$ with start offset $s$ and end offset $e$ consists of the contiguous values of $T$ from position $s$ to position $e$, i.e., $T_{s,e} = (t_{\tau-d+s},\dots,t_{\tau-d+e})$ with $1\leq s \leq e \leq d$. The length of $T_{s,e}$ is $|T_{s,e}| = e-s+1$.
\end{definition}

We refer to the length of subsequences as \emph{width}. Figure~\ref{fig:stream_example} illustrates subsequences in a sliding window. Periodic TS streams generally repeat a subsequence of values after a constant period of time, which we refer to as a temporal pattern (or period). However, periods can vary or drift, and local parts of TS may differ in terms of period length, shape or amplitude.

\begin{definition}
A \emph{segmentation} of a TS stream $S$ produces the latest completed segment of $S$ as a variable-sized interval $s_{-1} = [t_{c_{-2}}, \dots, t_{c_{-1}}]$ where $c_{-2} < c_{-1} \leq \tau$ are the two last discovered change points (or splits). For consistency, we consider the first observed value from $S$ as the first change point.
\end{definition}

The location and amount of CPs in $S$ is unknown and must be inferred by evaluating the last $d$ data points in the sliding window $S_{\tau-d+1,\tau}$. Change points and the respective segments are continuously reported until $S$ is aborted. Note that, by definition, the latest reported segment may stretch until before the current window.

\begin{definition}
The problem of streaming time series segmentation (STSS) is to find a meaningful segmentation of a given TS stream $S$ such that the change points between two subsequent segments correspond to state changes in the observed process. 
\end{definition}

The notion of being \emph{meaningful} depends on the domain, typically relating to the shape or value distribution of potential segments. Following~\cite{gharghabi2017matrix}, we assume that a natural process has discrete states that lead to changes in measured values. An example are sequences of human emotional states, that can switch between (a) resting, (b) amused or (c) stressed and influence biosignals, as studied in~\cite{Schmidt2018IntroducingWA}. The task of STSS would be to track e.g. the last 10 seconds of a subject's respiration signal and report the last completed segment (e.g. resting), as soon as another one (e.g. stressed) emerges. This differs from other problems such as trend detection~\cite{Truong2019SelectiveRO}. 

STSS algorithms must maintain efficient data structures with constant memory requirements and minimal latency to be able to report segmentations in real-time. They must also possess the ability to decide when they have seen enough data points to predict a CP, using only the limited information available from $S_{\tau-d+1,\tau}$.




\subsection{Self-supervised Time Series Segmentation}

In order to detect the last completed segment, we must be able to differentiate it from the newly evolving one. ClaSS enumerates potential binary segment candidates (splits of $S_{\tau-d+1,\tau}$ into two parts). We assess the distinctiveness between candidates using a heterogeneity score, selecting the segmentation with the most dissimilar segments, provided the score surpasses a predefined threshold.

Our scoring methodology draws inspiration from self-supervised change analysis, as formulized by Hido et al.~\cite{Hido2008UnsupervisedCA}. In self-supervised learning, an unsupervised learning variant, the data itself generates supervision labels. Consider two data sets, $X_A$ and $X_B$, representing subsequences from different segment candidates. We assign label $0$ to $X_A$ instances and $1$ to $X_B$ instances, facilitating a binary classification evaluation using cross-validation. A TS classifier, trained on labelled subsequences, predicts labels for unlabelled instances. We employ cross-validation, wherein the classifier is trained on $(k-1)$ portions of the data and tested on the remaining part. A $k$-NN classifier, for instance, collects and aggregates subsequence labels from train instances. This process repeats $k$ times, covering all combinations, with the average of the $k$ evaluation scores (e.g., F1-scores) representing the classifier's performance. This value measures the classifier's ability to distinguish between data sets $X_A$ and $X_B$. A high score implies high dissimilarity and unique characteristics between the segments, while a lower score indicates similarities, suggesting the data sets may belong to the same segment.

\subsection{Classification Score Profile}

ClaSS is based on the idea of the self-supervised TSS algorithm ''Classification Score Profile'' (ClaSP), as introduced for the batch setting in~\cite{Schfer2021ClaSPT}. We briefly recapitulate the concept of ClaSP. In Section 3, we describe how ClaSS efficiently computes ClaSP to address the streaming case.

\begin{definition}
	Given a TS $T$, $|T| = n$ and a subsequence width $w$, a ClaSP is a real-valued sequence of length $n-w+1$, in which the $i$-th value is the cross-validation score $c \in [0,1]$ of a classifier trained on a binary classification problem with overlapping labelled subsequences [($T_{1,w}, \textbf{0}$), \dots, ($T_{i-w+1,i}, \textbf{0}$), ($T_{i-w+2,i+1}, \textbf{1}$), \dots ($T_{n-w+1,n}, \textbf{1}$)], with labels $0$ and $1$.
\end{definition}

Conceptually, a ClaSP is the result of a sequence of self-supervised TS classifications, summarized in a profile that annotates $T$, where every offset (or split point) $i$ reports how well a TS classifier can differentiate the left from the right subsequences (see Figure~\ref{fig:class_example} bottom). ClaSP quantifies the heterogeneity between potential segments as a profile.

The main drawback of ClaSP is its high runtime complexity of $\mathcal{O}(n^2)$. Directly applying it for the streaming case on high-frequency streams with a sliding window (of length $d$) is impracticable, as it requires $\mathcal{O}(d^2)$ computations for each new observation in the TS stream $S$. Furthermore, such an approach would force the method to take decisions on CPs only based on the current sliding window, which leads to false positives.

\section{Classification Score Stream} \label{sec:class}

\begin{figure}[t]
	\includegraphics[width=1.0\columnwidth]{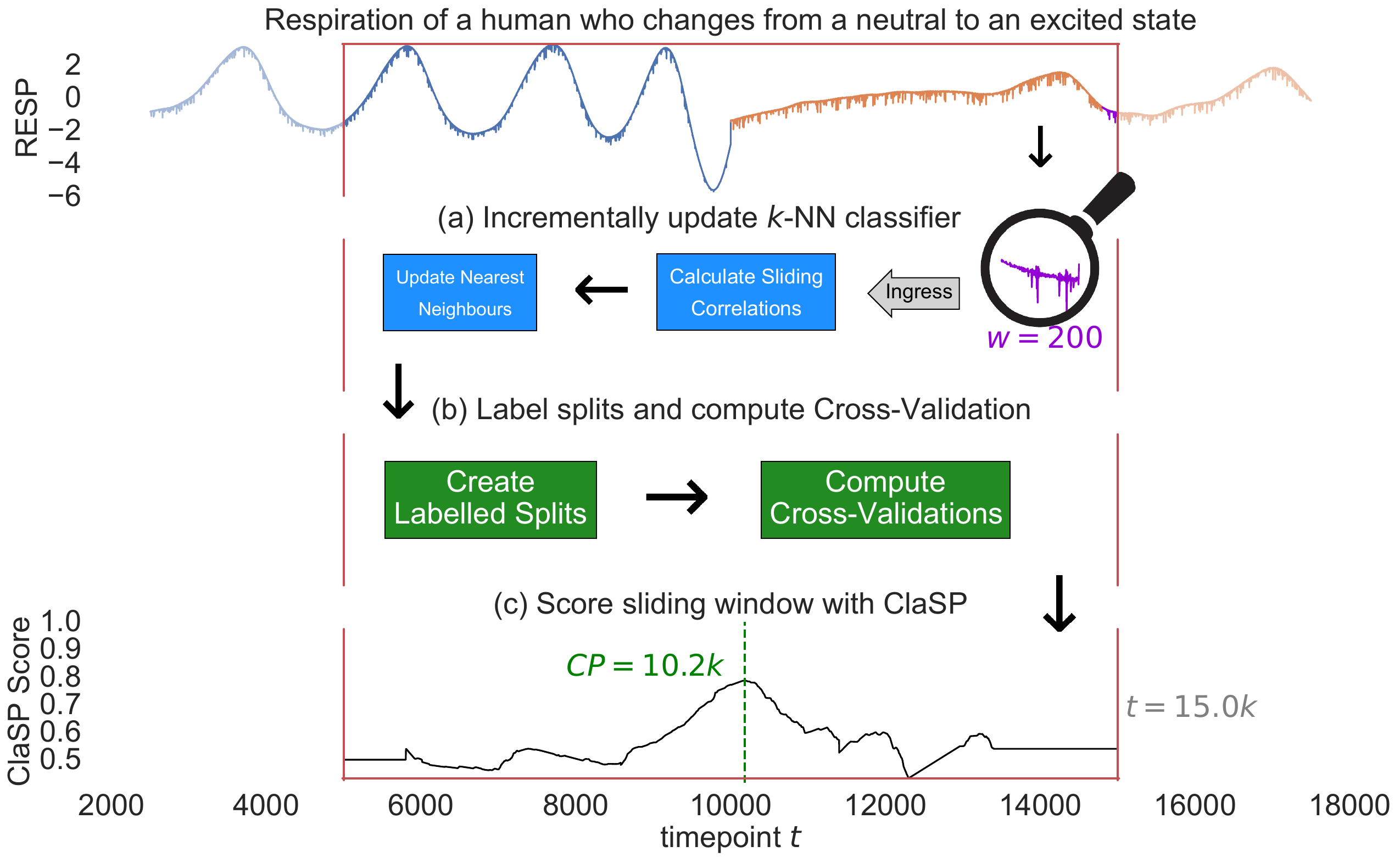}
	\caption{The conceptual ClaSS workflow for a human respiration recording that captures the transition from a neutral to an excited state~\cite{Schmidt2018IntroducingWA}. (a) The streaming $k$-NN classifier in ClaSS is updated with the newest subsequence (magenta). (b) For every possible offset, the sliding window (red) is transformed into hypothetical binary classification problems evaluated using cross-validation. (c) The result, ClaSP, annotates the sliding window.\label{fig:class_workflow}
	}
\end{figure}

We propose \emph{Classification Score Stream (ClaSS)}, a novel method for fast and accurate STSS. ClaSS uses a sliding window to update a streaming $k$-nearest neighbour ($k$-NN) classifier (with correlation as similarity measure) from continuous TS streams, computing the homogeneity of hypothetical sliding window splits and applying hypothesis testing to determine statistically significant CPs. A high-level overview of the workflow of ClaSS is illustrated in Figure~\ref{fig:class_workflow} and presented by pseudocode in Algorithm~\ref{alg:class}.

\begin{algorithm}[t]
	\caption{Classification Score Stream}\label{alg:class}
	\begin{algorithmic}[1]
        \Statex $this.N \gets $ array of length $(d-w+1) \times 3$  \Comment{$k$-NN indices}
        \Statex $this.C \gets $ array of length $(d-w+1) \times 3$  \Comment{$k$-NN correlations}
		\Procedure{class}{$S$, $d$}		  
			\State $cp_l \gets d$
			\State $w \gets \textsc{learn\_subsequence\_width}(S, d)$
			
			\While{\textsc{has\_next}($S$)}
                \State $S_{\tau-d+1,\tau} \gets $ retrieve last $d$ time points from $S$
                \State $cp_l \gets $ \textsc{max}($1,cp_l-1$) \Comment{Account for shift in $S_{\tau-d+1,\tau}$}
                \State $\textsc{update\_streaming\_knn}(this.C, this.N, S_{\tau-d+1,\tau}, w, 3)$
			      \State $\textsc{ClaSP} \gets \textsc{cross\_val\_scores}(this.N_{cp_l,d-w+1}, w)$
                
                \If{\textsc{has\_significant\_CP}(\textsc{ClaSP})}
                    \State $cp \gets cp_l + \textsc{argmax}(\textsc{ClaSP})-1$
                    \State $\textsc{report}(\tau - d + cp)$
                    \State $cp_l \gets cp$
                \EndIf
            \EndWhile
		\EndProcedure
	\end{algorithmic}
\end{algorithm}

The method takes a time series stream $S$ and the size of the sliding window $d$ as inputs, and first learns a subsequence width $w$ as a model-parameter from the first $d$ observations in the stream (line~3, see Subsection~\ref{sec:model-param}). It then processes a single observation from $S$ at a time (line~4). The procedure stores the last $d$ data points, its sliding window, in the sequence $S_{\tau-d+1,\tau}$ (line~5) and updates the position of the last CP $cp_l$ that denotes the beginning index of the yet unsegmented values (line~6). For consistency reasons, we consider the first observed value from $S$ as the first CP. Note that $S_{\tau-d+1,\tau}$ contains a prefix of $d-1$ known data points while only the last measurement is new. We exploit this property later to speed up computation. ClaSS maintains a 2-dimensional $k$-NN sliding window profile $N$ and pairwise Pearson correlations $C$ (line~7). $N$ maps the $i$-th subsequence $T_{i,i+w-1}$ to its $k$-NN subsequences in $S_{\tau-d+1,\tau}$. $C$ stores the pairwise correlations between a subsequence and its $k$-NNs. ClaSS computes the $d-cp_l-w+2$ cross-validation scores for the most recent unsegmented observations in $S$ (line~8), leaving only the newest $w-1$ observations unscored. This scoring process enables the method to determine the homogeneity of all hypothetical splits since the last CP $cp_l$. This step is also the most time-consuming component of the algorithm. Every local maximum in this profile marks a potential CP, as it distinguishes between the TS parts to its left and right with high accuracy. ClaSS checks for a significant CP $cp$, and if so, immediately reports its time point $\tau - d + cp$ to the user, resetting the last CP $cp_l$ (lines~9--13). The last discovered segment is then easily extracted with the last and current CP. The applied hypothesis testing is conservative in its predictions and reports CPs with high accuracy. The segmentation is then repeated as long as $S$ produces new observations. 

In the following subsections, we provide the details of the most important steps performed by ClaSS. We describe in Subsections \ref{sec:update-knn} and~\ref{sec:label-gt} how to update and evaluate the $k$-NN classifier as new observations arrive, \ref{sec:cpd} shows how we apply statistical tests to filter for false positive CPs, \ref{sec:model-param} explains how ClaSS automatically learns $w$, and~\ref{sec:hyper-param} discusses the setting of the hyper-parameter $d$. Finally, Subsection~\ref{sec:complexity} analyses the runtime and space complexity of ClaSS.


\subsection{Streaming $k$-Nearest Neighbours} \label{sec:update-knn}

In an offline setting, the computation of the $k$-NN profile can be delegated to a pre-processing step, which is then used for scoring hypothetical splits of the TS. This computation can be efficiently executed using various exact or approximate optimization techniques~\cite{Zhu2017ExploitingAN,SchallZimmerman2019MatrixPX}, GPUs~\cite{Zhu2016MatrixPI}, or index structures~\cite{Levchenko2020BestNeighborEE}. However, in a streaming context, such pre-processing becomes challenging as $S_{\tau-d+1,\tau}$ and $k$-NNs evolve with each time step. Upon the arrival of a new data point, it is ingressed into the sliding window $S_{\tau-d+1,\tau}$ at position $d$, while all preceding data points shift left by one, possibly egressing the oldest observation. The prevalent optimization techniques fall short in accommodating sliding windows, or require a-priori construction of prediction models or complex data structures, necessitating continuous updates.

\begin{figure}[t]
	\includegraphics[width=1.0\columnwidth]{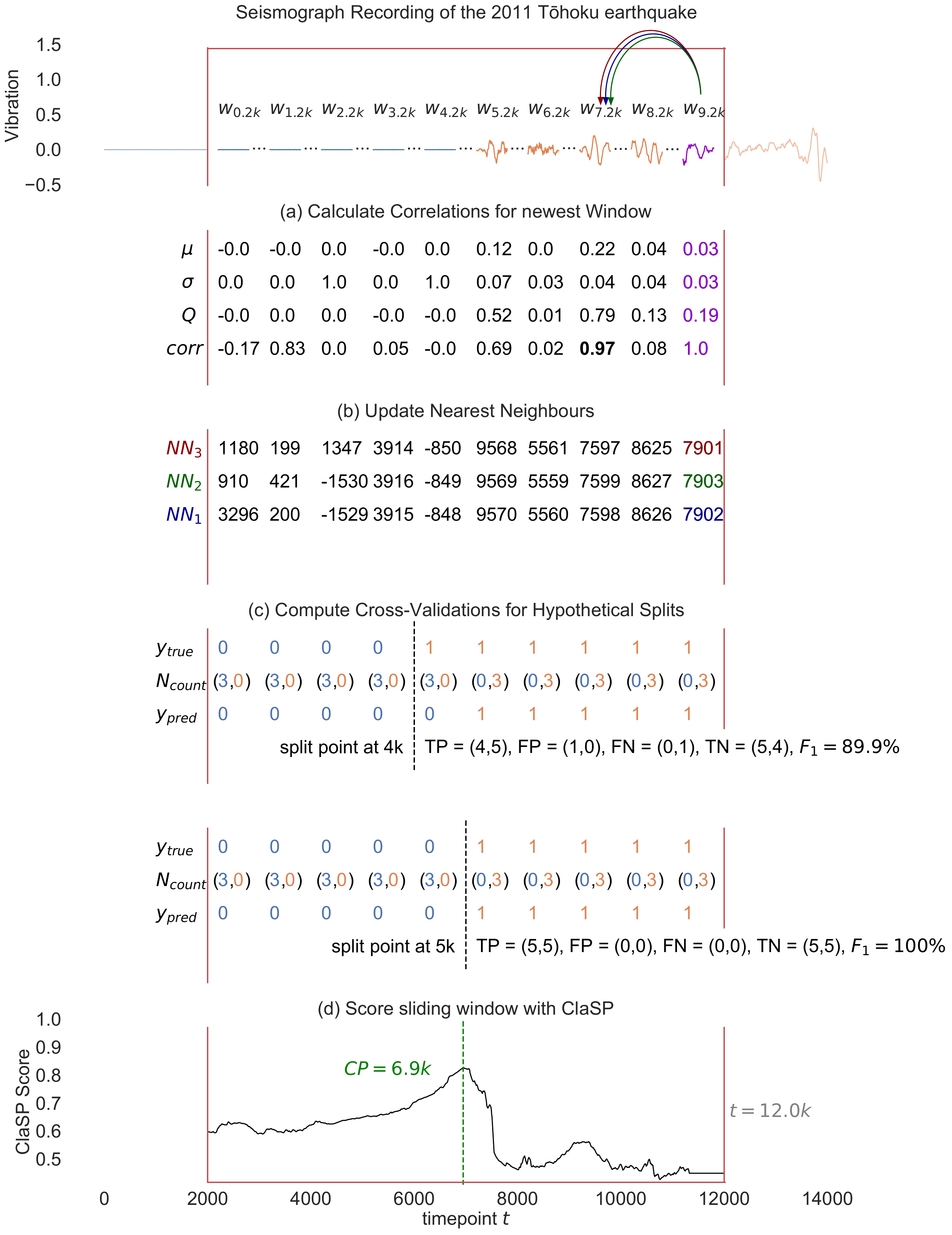}
	\caption{A workflow example for the $k$-NN classifier update and cross-validation computation in ClaSS. The TS stream contains the beginning of the 2011 Tōhoku earthquake seismogram, captured at Black Forest Observatory~\cite{Beyreuther2010ObsPyAP}. (a) The streaming 3-NN updates its means, standard deviations, and dot products to calculate the correlations between the latest subsequence (magenta) and the previous ones. (b) The 3-NN correlations and offsets are updated with the three highest correlations and their locations. (c) The sliding window is repeatedly divided into hypothetical splits and the updated $k$-NN classifier is evaluated to calculate the resulting classification scores (d) that form the ClaSP.  
	\label{fig:knn_scoring_workflow}
	}
\end{figure}

We propose the first exact streaming TS $k$-NN algorithm that runs in $\mathcal{O}(k \cdot d)$, substantially improving upon the fastest exact algorithm that requires $\mathcal{O}((k + \log d) \cdot d)$~\cite{Gharghabi2018DomainAO}. The central idea is to establish and maintain data structures, as the stream evolves, to incrementally compute $k$-NNs and update $C$ and $N$ accordingly. To do so, we have to perform three key operations: (a) calculate and store the $k$-NNs for the current (latest) subsequence to insert it into the data structure; (b) shift the existing $k$-NNs left and deal with out-of-range references that point out of the window; and (c) update the outdated existing $k$-NNs that may now point to the current subsequence. We first describe the mathematics of the similarity measure computation used for $k$-NN determination, and then specify how to efficiently implement steps (a) to (c) in Algorithm~\ref{alg:knn}. The workflow is visualized in Figure~\ref{fig:knn_scoring_workflow} (a--b).

\paragraph{Similarity Calculation} For every incoming data point, we determine the $k$-NNs between the newest subsequence $T_{d-w+1,d}$ and the maximal $d-w+1$ many subsequences (of size $w$) in $T = S_{\tau-d+1,\tau}$ by calculating their mutual correlations. This can be naively computed in $\mathcal{O}(d \cdot w)$ or optimized using Fast Fourier Transform (FFT) in $\mathcal{O}(d \log d)$~\cite{Yeh2016MatrixPI}, which is the basis of~\cite{Gharghabi2018DomainAO}. However, we can further improve the efficiency of this computation to $\mathcal{O}(d)$ by adapting ideas from the STOMP algorithm~\cite{Zhu2017ExploitingAN} to the streaming setting. The Pearson correlation $c^w_{i,j}$ (Equation~4) between two $w$-length subsequences starting at offset $i$ and $j$ in $T$ can be re-written using the dot product $q^w_{i,j}$~\cite{Mueen2014TimeSJ}. This definition mainly depends on the $w$-length subsequence means $\mu^w$, standard deviations $\sigma^w$ and dot products $q^w$. Rakthanmanon et al.~\cite{Rakthanmanon2012SearchingAM} showed that $\mu^w_l$ and $\sigma^w_l$ can be computed in $\mathcal{O}(1)$ from $\mu^w_{l-1}$ and $\sigma^w_{l-1}$ (independent of $w$) using so-called differencing cumulative running sums (Equation~1 and~2). 

\begin{align}
    \mu^w_l &= \frac{1}{w} \cdot \left(\textsc{cumsum}(T_{1,l+w-1}) - \textsc{cumsum}(T_{1,l-1})\right) \\ 
    \sigma^w_l &= \sqrt{\frac{1}{w} \cdot \left(\textsc{cumsum}^2(T_{1,l+w-1}) - \textsc{cumsum}^2(T_{1,l-1})\right) - (\mu^w_l)^2}
\end{align}

Furthermore, Zhu et al.~\cite{Zhu2017ExploitingAN} demonstrated that the dot product $q^w_{i,j}$ can be calculated in $\mathcal{O}(1)$ from $q^{w}_{i-1,j-1}$ (also independent of $w$) by reusing dot products from previous subsequences. Utilizing these two findings from batch analysis, we establish them for the streaming setting to compute the Pearson correlations between the newest subsequence $T_{d-w+1,d}$ and its $d-w$ predecessors in $\mathcal{O}(d)$ based on accessible information from the previous update.

\begin{align}
    q^w_{i,j} &= q^{(w-1)}_{i,j} + T_{i+w-1} \cdot T_{j+w-1} \\
    c^w_{i,j} &= \frac{q^w_{i,j} - w \mu^w_i \mu^w_j}{w \sigma^w_i \sigma^w_j} \\
    q^{(w-1)}_{i,j} &= q^w_{i-1,j-1} - T_{i-1} \cdot T_{j-1}
    \label{eq:correlation}
\end{align}

To do so, we first compute the means $\mu^w$ and standard deviations $\sigma^w$ from differenced (squared) running sum sliding windows. We then reuse the dot products between the $(w-1)$-length subsequences $T_{i,i+w-2}$ and $T_{d-w+1,d-1}$ from the last update, and add $T_{i+w-1} \cdot T_d$ to obtain the $w$-length dot products needed for the current iteration (Equation~3). Using the means, standard deviations, and dot products, we calculate the correlations (Equation~4) to determine the $k$-NNs for the current subsequence $T_{d-w+1,d}$. We then subtract $T_{i} \cdot T_{d-w+1}$ from the dot products to prepare them for the next update (Equation~5). Keeping the dot products updated, enables us to continuously reuse them to calculate correlations.

The similarity measure used in the streaming $k$-NN is not necessarily restricted to Pearson correlation; it can easily be adapted to (dis-)similarity functions that can be expressed with dot products, such as (complexity-invariant) Euclidean distance~\cite{Batista2013CIDAE}. We implement multiple measures that cover different stream properties.

\begin{algorithm}[t]
	\caption{Streaming $k$-Nearest Neighbors}\label{alg:knn}
	\begin{algorithmic}[1]    
        \Statex $this.R$ =  array of length $d$ \Comment{Cumsums}
        \Statex $this.R^2$ = array of length $d$ \Comment{sqrd. Cumsums}
        \Statex $this.Q$ = array of length $d-w+1$ \Comment{Dot products}
		\Procedure{calc\_knn}{$S_{\tau-d+1,\tau}$, $w$, $k$}		
            \State $T, start, end \gets S_{\tau-d+1,\tau}, d - \textsc{length}(S_{\tau-d+1,\tau}) + 1, d-w+1$

            \State \textbf{if} $\textsc{length}(T) < w$ \textbf{then} \textbf{return} \text{null, null} \textbf{end if}
		    
		    \State $\mu \gets \textsc{mean}(this.R) $ \Comment{Eqn. 1} 
            \State $\sigma \gets \textsc{std}(this.R, this.R^2) $ \Comment{Eqn. 2}
		    
		    \If{$start > 1$}
		        \State $this.Q_{start} \gets \textsc{dot}(T_{start,start+w-2}, T_{end,d-1})$
		    \EndIf

            \State add $T_{start+w-1,d} \cdot T_d$ to $this.Q_{start,end}$ \Comment{Eqn. 3}
		    \State $corr \gets \textsc{pearson}(this.Q, w, \mu, \sigma)$ \Comment{Eqn. 4}
		    \State $knn \gets \textsc{argkmax}(corr, k, w)$
		    \State subtract $T_{start,end} \cdot T_{end}$ from $this.Q_{start,end}$ \Comment{Eqn. 5}
		    \State\Return{$corr, knn$}
		\EndProcedure
        \Statex 
		
		\Procedure{update\_streaming\_knn}{$C$, $N$, $S_{\tau-d+1,\tau}$, $w$, $k$}	
	    \State $\textsc{shift\_add\_last}(this.R, this.R_d + S_{-1})$  
        \State $\textsc{shift\_add\_last}(this.R^2, this.R_d^2 + S_{-1}^2)$ 
        \State $corr, knn \gets \textsc{calc\_knn}(S_{\tau-d+1,\tau}, w, k)$
        
	    \State \textbf{if} $\textsc{length}(S_{\tau-d+1,\tau}) < w+k$ \textbf{then} \textbf{return} \textbf{end if}
	    \State $\textsc{shift\_add\_last}(C, corr[knn]), \textsc{shift\_add\_last}(N, knn)$
	    
	    \State $N_{1,d-w} \gets N_{1,d-w}-1$

        \State $mask \gets \textsc{changed\_nn\_pos}(C, N, corr, knn)$
        \State $\textsc{update}(C, mask, corr), \textsc{update}(N, mask, d-w+1)$
		\EndProcedure
	\end{algorithmic}
\end{algorithm}

\textbf{Incremental $k$-NN Calculation:} In the batch setting, we can calculate initial dot products for all subsequences using FFT~\cite{Yeh2016MatrixPI} and then update them to calculate $k$-NNs~\cite{Zhu2017ExploitingAN}. This preprocessing is not possible in the streaming setting, where we need to compute and update $k$-NNs as soon as new data points arrive and old ones are evicted. Therefore, we first enlarge the dot products and incrementally update them as outlined.  

Algorithm~\ref{alg:knn} takes the $k$-NN correlations $C$ and $k$-NN indices $N$, the sliding window $S_{\tau-d+1,\tau}$, the subsequence width $w$ and number of neighbours $k$ as input from Algorithm~\ref{alg:class}. It maintains two (squared) running sum sliding windows $R$ (or rather $R^2$) as class variables, which it updates and uses to calculate the Pearson correlation efficiently (lines~16--17). Subsequently, the algorithm calculates the correlations between the newest subsequence $T_{d-w+1,d}$ and the maximal $d-w+1$ subsequences (of size $w$) in $S_{\tau-d+1,\tau}$ to identify its $k$-NNs. The \textsc{calc\_knn} subroutine (lines~1--14) first computes the \emph{start} and \emph{end} index of the data contained in the sliding window (line~2) and then calculates the $d-w+1$ means $\mu$ and standard deviations $\sigma$ with $R$ (or rather $R^2$) (lines~4--5). Similarly, the procedure maintains $d-w+1$ dot products between the $(w-1)$-length subsequences $T_{i,i+w-2}$ and $T_{d-w+1,d-1}$ at the $i$-th offset as a sliding window class variable $Q$. For the first $d$ data points, it continuously enlarges $Q$ to include the correct $(w-1)$-length dot products (lines~6--8). The algorithm adds $T_{start+w-1,d} \cdot T_d$ to $Q_{start,end}$ to obtain $w$-length dot products (line~9). It then calculates the correlations between the newest and all other subsequences and determines its nearest neighbours with $k$ sequential searches, considering an exclusion radius of the last $\frac{3}{2}w$ observations to avoid trivial matches (lines~10--11). Lastly, the subroutine subtracts $T_{start,end} \cdot T_{end}$ from $Q$ to restore the $(w-1)$-length dot products for the next update and reports the correlations and $k$-NNs (lines~12--13) to \textsc{update\_streaming\_knn}. Being able to provide the dot products in every iteration, is the central optimization that leads to linear runtime, as opposed to recomputing them in log-linear time~\cite{Gharghabi2018DomainAO}. 

Figure~\ref{fig:knn_scoring_workflow} (a) shows an example of this process. In the last column, the procedure stores the mean and standard deviation for the newest subsequence (magenta) and in the last two rows its pairwise dot products and correlations with all previous subsequences. 

\textbf{$k$-NN Shift:} Having the $k$-NNs of the newest subsequence calculated, the procedure updates the correlations $C$ and offsets $N$ for the newest subsequence $T_{d-w+1,d}$ (line~20) and shifts the existing ones left accordingly. This leaves the prior $d-w$ to be adjusted. The algorithm decreases their offsets in $N$ by one, to account for the shift (line~21), potentially producing negative out-of-range indices that point to subsequences outside the sliding window. To avoid this, we could constrain the nearest neighbour direction, as proposed in~\cite{Gharghabi2018DomainAO}. However, for the $k$-NN classifier in ClaSS, we do not even need the actual subsequences, but only their offsets, which by design belong to class zero if they are negative. Thus, we may safely ignore this issue. 


\textbf{$k$-NN Update:} Lastly, the algorithm checks if the newest subsequence is one of the $k$-NNs of the existing subsequences in $T$. If so, the correlations and offsets are updated (lines~22--23). This can be done efficiently by locating the offsets that have $k$-NNs with lower or equal correlations compared to the newest subsequence, and inserting it in order of descending correlation, while expelling the least correlated one. Subsequently, the correlations $C$ and offsets $N$ are updated with the new observation. 

Figure~\ref{fig:knn_scoring_workflow} (b) illustrates an example of the updated offsets in $N$. The last column contains the 3-NN for the newest subsequence, and it can now be a NN of previous subsequences.

\subsection{Scoring the Sliding Window} \label{sec:label-gt}

The basic idea of self-supervised learning for obtaining scores for hypothetical split points is to first assign artificial ground truth labels to each subsequence up to (after) a split point, assigning them to class zero (one). These labels are then used in a second step to create the predictions of the classifier for every subsequence by the $k$-NN rule, collecting and aggregating ground truth classes into majority labels. In a third and final step, we calculate a classification score with both the ground truth and predicted labels. This calculation is repeated for all possible splits, thus creating the classification score profile (ClaSP). 

For the at most $d-w+1$ subsequences in $N$, the implementation in \cite{Ermshaus2022ClaSP} computes a single classification score in $\mathcal{O}(d)$, re-using the class-independent $k$-NN offsets and relabelling them according to the changing ground truth labels. However, since this cross-validation is executed $(d - 2 \cdot w - 1)$ times, it becomes inefficient in the streaming setting, resulting in $\mathcal{O}(d^2)$ computations at the arrival of a single data point. 

We propose a novel algorithm for cross-validating a self-supervised $k$-NN classifier that runs in $\mathcal{O}(d)$ time, exploiting the observation that the label configurations for two consecutive splits only minimally differ. It computes this delta in amortized constant runtime, as opposed to creating new labels for each cross-validation, substantially accelerating the process while being exact. This key idea is implemented in Algorithm~\ref{alg:cross_val}, and visualized in Figure~\ref{fig:knn_scoring_workflow} (c--d).


Algorithm~\ref{alg:cross_val} receives the sliding window $k$-NN indices $N$ and the subsequence width $w$ as input from Algorithm~\ref{alg:class}. It initializes the ground truth and predicted labels $y_{true}$ and $y_{pred}$ as two arrays of $d - w + 1$ ones, and stores the label counts for each subsequence in a 2-dimensional array $N_{count}$ (line~2) of size $(d-w+1) \times 2$. This data structure stores for the $i$-th subsequence $T_{i,i+w-1}$ the number of 0 and 1 labels within its $k$-NN. The procedure transposes the $k$-NN indices $N$ to its reverse NN $R$, which retrieves all subsequences that have a given offset as their $k$-NN (line~3). This information is necessary for retrieving the relevant offsets during the relabelling process. The algorithm also initializes a confusion matrix to store the true positive (TP), false positive (FP), false negative (FN) and true negative (TN) counts for both labels (line~4). For the 0 class, all measures are initialized to 0 except the TN count, which is $d-w+1$. Conversely for the 1 class, the TP count is $d-w+1$ with all other counts being 0. The confusion matrix is updated between cross-validations and used to calculate the classification scores for each split in constant time.

The relabelling procedure changes the ground truth label in $y_{true}$ from 1 to 0 for a given split index $i \in [w+1, \dots, d-w-1]$ (line~7), updating the relevant $k$-NN labels and the confusion matrix in the process (line~8-12). The classification score resulting from this is stored in the ClaSP (line~13). To achieve this efficiently, the procedure retrieves the subsequence offsets with $R$ which have the split index $i$ as one of their $k$-nearest neighbours (line~8). The count of zero (one) labels is increased (decreased) by one to reflect that the ground truth label changed from 1 to 0 at split point $i$ (line~9). The algorithm then computes their predicted $k$-NN majority label, updating the predicted labels $y_{pred}$ and confusion matrix $M$ accordingly (line~10--11). This is done by subtracting (adding) the old (new) prediction from (to) $M$, and replacing it in $y_{pred}$. The classification score for split point $i$ is computed and stored in ClaSP (line~13). Evaluation scores such as accuracy and F1 can be calculated with $M$ in constant runtime. Lastly, the cross-validation scores are returned, which constitute ClaSP for the current sliding window in ClaSS (line~15). The central optimization in this routine exploits the fact that although a subsequence can be a $k$-NN to many other subsequences, the total number of nearest neighbours is bound by exactly $k \cdot (d-w+1)$, the size of all lists from $R$.

Figure~\ref{fig:knn_scoring_workflow} (c) exemplifies how the label configuration changes from split point $i$ to $i+1$. At this offset, the ground truth label changes from 1 to 0. For all offsets that have the split point as a NN, the procedure updates the label counts, confusion matrix and computes new predictions. Lastly, it calculates the classification score for the split point that is inserted in ClaSP (d) at offset $i+1$.

\begin{algorithm}[t]
	\caption{Cross Validation Scores}\label{alg:cross_val}
	\begin{algorithmic}[1]
		\Procedure{cross\_val\_scores}{$N$, $w$}		
		    \State $N_{count}, y_{true}, y_{pred} \gets \textsc{init\_labels}(N)$
		    \State $R \gets \textsc{transpose}(N)$ \Comment{Reverse NN}
		    \State $M \gets \textsc{init\_conf\_matrix}(y_{true}, y_{pred})$ 
		    \State $ \textsc{ClaSP} \gets \text{ initialize array of length } d$ 
		    
		    \For{$i \in [w+1, \dots, d-w-1]$}
		        \State $y_{true}[i-w] \gets 0$ \Comment{The updated label}
                \For{$idx \in R[i-w]$} \Comment{Affected NN}
                    \State $zeros, ones \gets \textsc{update\_counts}(N_{count}[idx])$
                    \State $y_{pred}[idx] \gets 0 \text{ if } zeros \geq ones \text{ else } 1$
                    \State $\text{update } M \text{ with } y_{pred}[idx]$
                \EndFor
            \State $\textsc{ClaSP}[i] \gets \textsc{score\_function}(M)$
		    \EndFor
		    \State \Return{$\textsc{ClaSP}$}
		\EndProcedure
	\end{algorithmic}
\end{algorithm}
 
\subsection{Detecting Significant Changes} \label{sec:cpd}

In principle, every local maximum in the cross-validation scores is a potential CP because it marks a sliding window split that separates two differently-shaped segments. This observation is useful for domain experts, who can use a visualization tool, such as~\cite{ClaSSWebpage}, to assess these points and to spot semantic changes in the incoming stream. However, automatic change point detection (CPD) is essential for stream segmentation to be incorporated as an IoT edge analytics tool~\cite{Krishnamurthi2020AnOO}, or to uncover latent segmentations in signals where no expert with domain knowledge is available. 


To implement this in ClaSS (Algorithm~\ref{alg:class}, line~9), we first locate the global maximum in the classification scores. We then use the non-parametric two-sided Wilcoxon rank-sum test, as suggested in~\cite{Ermshaus2022ClaSP}, to check whether, for the associated sliding window split $i$ (from Algorithm~\ref{alg:cross_val}, line~6), the difference in predicted label frequencies $y_{pred}$ after cross-validation between the left $y_{pred}[1 \dots i-w]$ and right $y_{pred}[i-w+1 \dots d-w+1]$ segment is likely due to chance or not. In the ablation study (Subsection~\ref{sec:design_choices}), we empirically learn a significance level for this test. However, in the streaming setting we run into the problem that the test statistics are calculated with different sample sizes due to the sliding window procedure, which takes $d$ as a hyper-parameter and only scores the most recent observations beginning at the last CP $cp_l$ (Algorithm~\ref{alg:class}, line~8). Accordingly, the number of the predicted cross-validation labels in ClaSP, with which the significance test is computed, is variable, resulting in a bias, because the p-value tends to decrease with increasing observations~\cite{Thiese2016PVI}. To control the variable sample size, resampling is used. $1k$ labels are randomly chosen with replacement from the cross-validation labels $y_{pred}$, maintaining the class distribution, in order to make the significance level independent of the sliding window size and increase accuracy.



\subsection{Learning the Subsequence Width} \label{sec:model-param}

Setting appropriate parameters is a crucial task for unsupervised data mining algorithms in general and for STSS in particular~\cite{Rijn2017HyperparameterIA}. Therefore, we propose methods to relieve users from this task and study the impact. A model-parameter in ClaSS is the subsequence width $w$ (Algorithm~\ref{alg:class}, line~3), needed to partition the TS stream into overlapping subsequences that can be classified. By default, we learn a proper value for $w$ on the first $d$ observations, under the assumption that these initial observations are representative of the characteristics of the entire stream. Multiple window size selection (WSS) methods have been developed based on the idea that a temporal pattern approximately repeats throughout a TS~\cite{Ermshaus2022WSS}, a presumption shared by ClaSS. We use the SuSS~\cite{Ermshaus2022ClaSP} algorithm for WSS, due to its expected linear (and worst-case log-linear) runtime complexity with respect to TS length. After the subsequence width has been determined at the start of ClaSS, the sliding window segmentation algorithm processes the stream from the first observation onward. 

In settings where users expect or encounter concept drifts in the TS stream, the subsequence width $w$ can be periodically relearned. Similar to the algorithm's initial phase, the data points from a newly evolving segment can be utilized to relearn $w$, and the segmentation process resumes. This behaviour can be activated on demand. Although we do not need to account for concept drifts in our experimental evaluation, it provides flexibility for applications where they are a concern.


\subsection{Setting the Sliding Window Size} \label{sec:hyper-param}

Like most streaming algorithms~\cite{Gama2007data}, ClaSS requires a sliding window size hyper-parameter $d$ (Algorithm~\ref{alg:class}, line~1). With larger values for $d$, ClaSS becomes more accurate, albeit slower, as the amount of available information increases. In many real-world data streams, this tradeoff exhibits a diminishing returns effect, where the accuracy of ClaSS initially improves as $d$ increases, but then tapers off for even larger values. This stagnation is expected, as the amount of information in a signal typically does not grow linearly with its size due to the presence of repetitive substructures~\cite{Ermshaus2022WSS}. Therefore, $d$ should be set to a value that covers multiple instances (10 to 100 times) of temporal patterns. If such knowledge is not available, ClaSS uses a default value of $10k$, which we found to be robust throughout many domains and sensor types in the experiments (see Subsection~\ref{sec:runtime}), leading to fast and accurate stream segmentations.

\subsection{Computational Complexity} \label{sec:complexity}

In a streaming setting, the runtime and space complexity of a segmentation procedure is of critical importance for its applicability, as it must keep up with real-time requirements. The complexity of ClaSS is mainly determined by the one-time subsequence width selection (Algorithm~\ref{alg:class}, line~3) and the recurring scoring and extraction of the sliding window (lines~7--9). SuSS requires $\mathcal{O}(d \log w)$ to learn the subsequence width from the initial $d$ observations.

The total runtime of the $k$-NN update is dominated by the dot product calculation in $\mathcal{O}(w)$ (Algorithm~\ref{alg:knn}, lines~6--8) and $k$ sequential NN searches in $\mathcal{O}(k \cdot d)$ (line~11). In the streaming setting, where the routine is called $n \gg d$ times, its time complexity is $\mathcal{O}(d)$, since $k$ is a small constant. The runtime of the shift operation (line~20) is dominated by moving the data, which is in $\mathcal{O}(k \cdot d) = \mathcal{O}(d)$ as both $C$ and $N$ have the dimensionality $(d-w+1) \times k$. Updating $C$ and $N$ (lines~22--23) relies on replacing and moving values, which is also performed in $\mathcal{O}(k \cdot d)$ and hence in $\mathcal{O}(d)$.

The complexity of the sliding window scoring depends on relabelling and evaluating $k$-NN offsets. For a single cross-validation, the amortized runtime is in $\mathcal{O}(1)$ (Algorithm~\ref{alg:cross_val}, lines~7--13), as only counts are updated and the score is calculated for a constrained number of offsets in the reverse NN. For all $(d - 2\cdot w - 1)$ splits, this leads to $\mathcal{O}(d)$ total runtime for the entire algorithm. 

The Wilcoxon rank-sum test in ClaSS, which is used to detect CPs (Algorithm~\ref{alg:class}, line~9), can also be implemented in $\mathcal{O}(d)$, as it mainly depends on ranking binary classes. This results in an overall amortized runtime complexity of $\mathcal{O}(d)$ for processing a single observation and $\mathcal{O}(n \cdot d)$ for segmenting $n$ measurements with ClaSS. The space complexity is likewise linearly dependent on the sliding window size.

\begin{table}[t]
    \begin{centering}
        \caption{Technical specifications of TS used for experiments.\label{tab:db_spec}}	
        \begin{tabular}{c|ccc}
            \toprule 			
            Name & No. & TS Length & No. Segments  \tabularnewline 
            & TS & Min/Median/Max & Min/Median/Max \tabularnewline \hline
            TSSB~\cite{Ermshaus2022ClaSP} & 75 & 240 / 3.5k / 20.7k & 1 / 3 / 9 \tabularnewline 
            UTSA~\cite{gharghabi2017matrix} & 32 & 2k / 12k / 40k & 2 / 2 / 3 \tabularnewline \hline
            mHealth~\cite{Baos2014mHealthDroidAN} & 90 & 32.2k / 34.3k / 35.5k & 12 / 12 / 12 \tabularnewline
            Arr DB~\cite{Moody2001TheIO} & 96 & 650k / 650k / 650k & 1 / 10 / 207 \tabularnewline
            VE DB~\cite{Greenwald1986TheDA} & 44 & 525k / 525k / 525k & 2 / 13 / 134 \tabularnewline
            PAMAP~\cite{Reiss2011TowardsGA} & 135 & 37.5k / 132.1k / 175k & 2 / 9 / 9 \tabularnewline
            Sleep DB~\cite{Kemp2000AnalysisOA} & 88 & 2.7M / 3.1M / 3.9M & 83 / 138 / 231 \tabularnewline
            WESAD~\cite{Schmidt2018IntroducingWA} & 32 & 2M / 2.1M / 2.1M & 5 / 5 / 5 \tabularnewline
            \bottomrule 			
        \end{tabular}
    \end{centering}    
\end{table}

\begin{table}[t]
    \begin{centering}
        \caption{Specification of competitors; $n$ is the number of all observed values, $c \ll d$ is an adaptive/custom window size.\label{tab:competitors}}	
        \begin{tabular}{c|cc}
            \toprule 			
            Competitor & Update Complexity & Segmentation Method  \tabularnewline \hline
            BOCD~\cite{adams2007bayesian} & $\mathcal{O}(n)$ & Bayesian probability \tabularnewline
            FLOSS~\cite{gharghabi2017matrix} & $\mathcal{O}(d \log d)$ & Matrix profile \tabularnewline
            ClaSS & $\mathcal{O}(d)$ & Self-supervision \tabularnewline
            ChangeFinder~\cite{Yamanishi2002AUF} & $\mathcal{O}(c^2)$ & Moving averages \tabularnewline
            Window~\cite{Truong2019SelectiveRO} & $\mathcal{O}(c)$ & Autoregressive cost \tabularnewline
            NEWMA~\cite{Keriven2018NEWMAAN} & $\mathcal{O}(c)$ & Moving averages \tabularnewline
            ADWIN~\cite{Bifet2007LearningFT} & $\mathcal{O}(\log c)$ & Adaptive Statistics \tabularnewline
            DDM~\cite{Gama2004LearningWD} & $\mathcal{O}(1)$ & Model error \tabularnewline
            HDDM~\cite{Blanco2015OnlineAN} & $\mathcal{O}(1)$ &  Hoeffding’s inequality \tabularnewline
            \bottomrule 			
        \end{tabular}
    \end{centering}    
\end{table}

\section{Experimental Evaluation} \label{sec:experiments}

To evaluate the characteristics of ClaSS and to compare it to 8 state-of-the-art competitors, we measured accuracy, runtime and scalability on large benchmark data sets as well as real-world annotated data archives from experimental studies. Subsection~\ref{sec:setup} outlines data sets, evaluation metrics and methods. We investigate the influence of different design choices in Subsection~\ref{sec:design_choices} through an ablation study. Subsections~\ref{sec:comparative_analysis} and~\ref{sec:runtime} further evaluate ClaSS and 8 competitors in terms of accuracy, runtime, and throughput. Lastly, Subsection~\ref{sec:usecases} discusses two real-life use cases to showcase the features and limitations of ClaSS. Experiments were conducted on an Intel Xeon 8358 with 2.60 GHz, 2 TB RAM, 128 cores, running Python 3.8. To ensure reproducibility and foster follow-up works, all source codes, Jupyter-Notebooks, TS used in the evaluation, visualizations, raw measurement sheets and a technical report are available on our supporting website~\cite{ClaSSWebpage}.

\subsection{Experiment Setup} \label{sec:setup}

\paragraph{Data Sets} We use $592$ time series from two public TSS benchmarks and six data archives from experimental studies (see Table~\ref{tab:db_spec}) to measure the quality of ClaSS and 8 competitors. In the following evaluations, we simulated the streaming setting by processing one data point at a time. The ground truth CP locations, used to evaluate the algorithms, were annotated by domain experts. The benchmark data sets consist of 107 preprocessed medium to large (240 to 40k) TS, representing a dense collection of diverse problem settings. The data archives contain 485 very large (32.2k to 3.9M) raw sensor signals from 10 sensors capturing human subjects in experimental studies. These data sets are of particular interest, as they reflect a common application of STSS, in which researchers must first segment instances of very large, heterogeneous recordings into homogeneous subsequences and then apply advanced data mining algorithms such as anomaly detection, forecasting, or classification. In detail, we use (see Table~\ref{tab:db_spec}):

\begin{enumerate}
    \item The UCR Time Series Semantic Segmentation Archive (UTSA)~\cite{gharghabi2017matrix} is a benchmark containing TS from the literature that capture the dynamics of biological, mechanical or synthetic processes. 
    
    \item The Time Series Segmentation Benchmark (TSSB)~\cite{Ermshaus2022ClaSP} features semi-synthetically created TS from the UCR archive~\cite{UCRClassification} from sensor, device, image, spectrogram and simulation signals. 

    \item PAMAP~\cite{Reiss2011TowardsGA} and mHealth~\cite{Baos2014mHealthDroidAN} contain multi-sensor ankle motion data of human subjects performing activities. 

    \item WESAD~\cite{Schmidt2018IntroducingWA} is an archive of physiological chest recordings from users in restful, excited and stressed states. 

    \item The Sleep DB~\cite{Kemp2000AnalysisOA} contains polysomnographic sleep recordings from multiple subjects capturing their sleep stages.

    \item MIT-BIH-VE DB~\cite{Greenwald1986TheDA} and MIT-BIH-Arr DB~\cite{Moody2001TheIO} feature ECG data from patients with normal cardiac activity, or sustained episodes of ventricular fibrillation or arrhythmias.
\end{enumerate}

Whilst the benchmarks (first two in the list) encompass a variety of domains, the data archives focus on human-centric processes.


\paragraph{Evaluation Metric}

The literature contains multiple classification- and clustering-based metrics to assess the quality of segmentations; see~\cite{Truong2019SelectiveRO} for a survey. Specifically, we use the soft evaluation measure Covering~\cite{van2020evaluation}. This measure quantifies the exact degree to which predicted vs annotated segments overlap and allows the comparison of different-sized (including empty) segmentations. 

It is defined as follows: Let the interval of successive CPs $[t_{c_i},\dots,t_{c_{i+1}}]$ denote a segment in $T$ and let $segs_{pred}$ as well as $segs_{T}$ be the sets of predicted or ground truth segmentations, respectively. For notational convenience, we always consider $t_{c_1} = 0$ as the first and $t_{c_{|segs_{T}|}} = n+1$ as the last CP to include the first (last) segment. The Covering score reports the best-scoring weighted overlap between a ground truth and a predicted segmentation (using the Jaccard index) as a normed value in the interval $[0,\dots,1]$ with higher being better (Equation~\ref{eqn:covering}). 
\begin{align}
    \textsc{Covering} = \frac{1}{\|T\|} \sum_{s \in segs_{T}} \|s\| \cdot \max_{s' \in segs_{pred}} \frac{\| s \cap s' \|}{\| s \cup s' \|}
    \label{eqn:covering}
\end{align}

\begin{figure}[t]
	\begin{minipage}{4cm}
        \includegraphics[width=1.0\columnwidth]{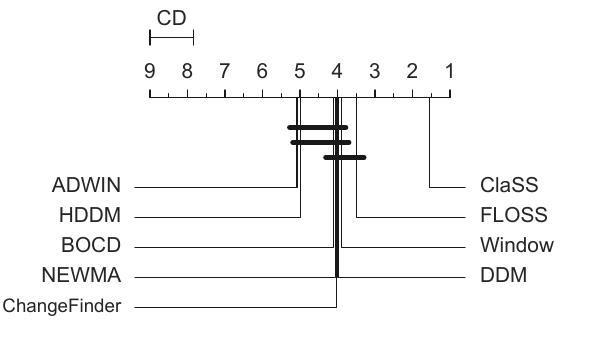}
	\end{minipage}
	\begin{minipage}{4cm}
        \includegraphics[width=1.0\columnwidth]{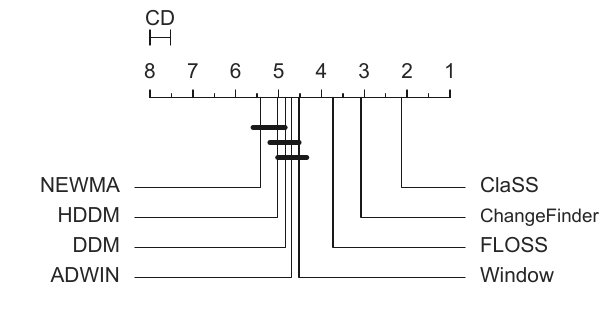}
	\end{minipage}

 	\begin{minipage}{4cm}
        \includegraphics[width=1.0\columnwidth]{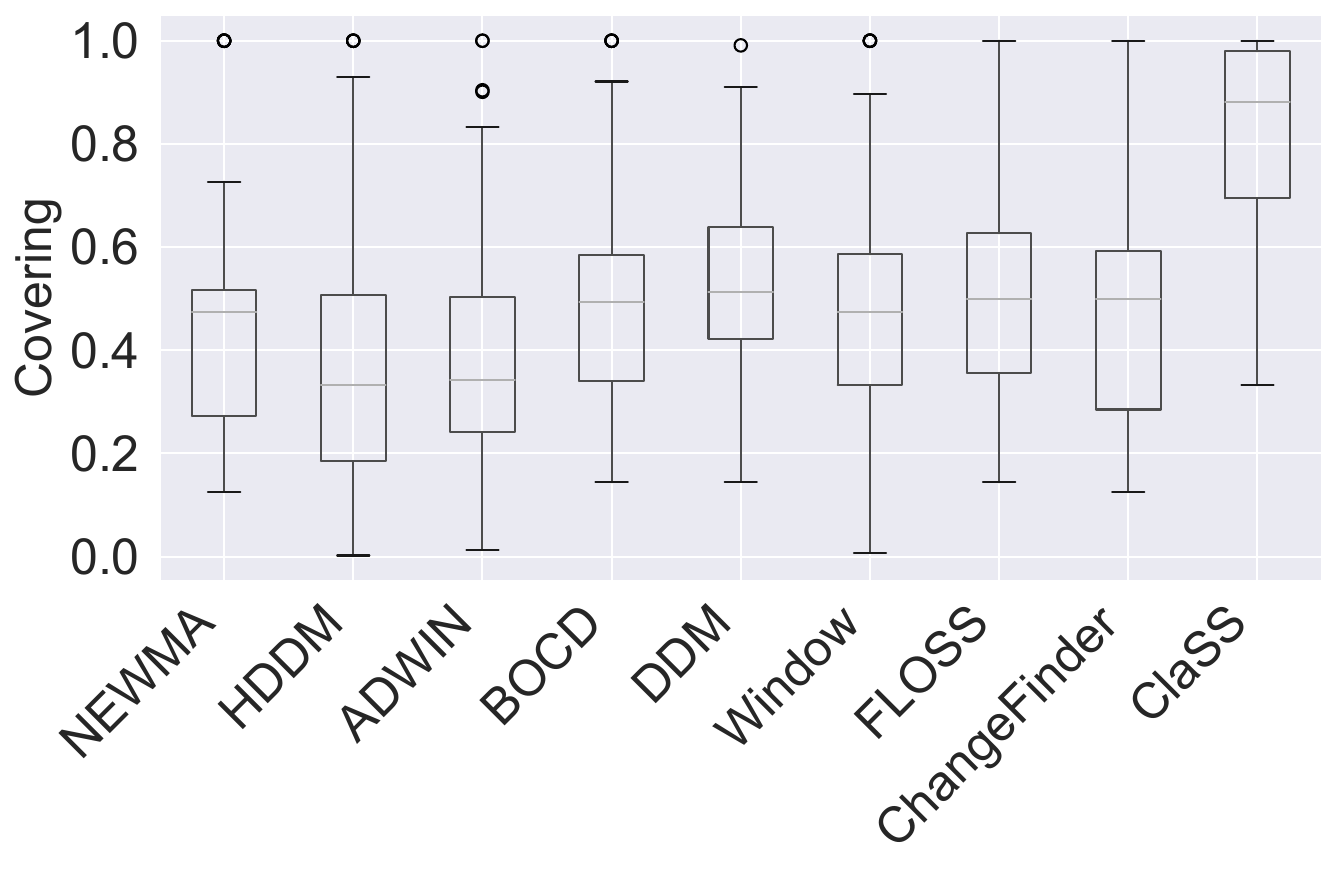}
	\end{minipage}
	\begin{minipage}{4cm}
        \includegraphics[width=1.0\columnwidth]{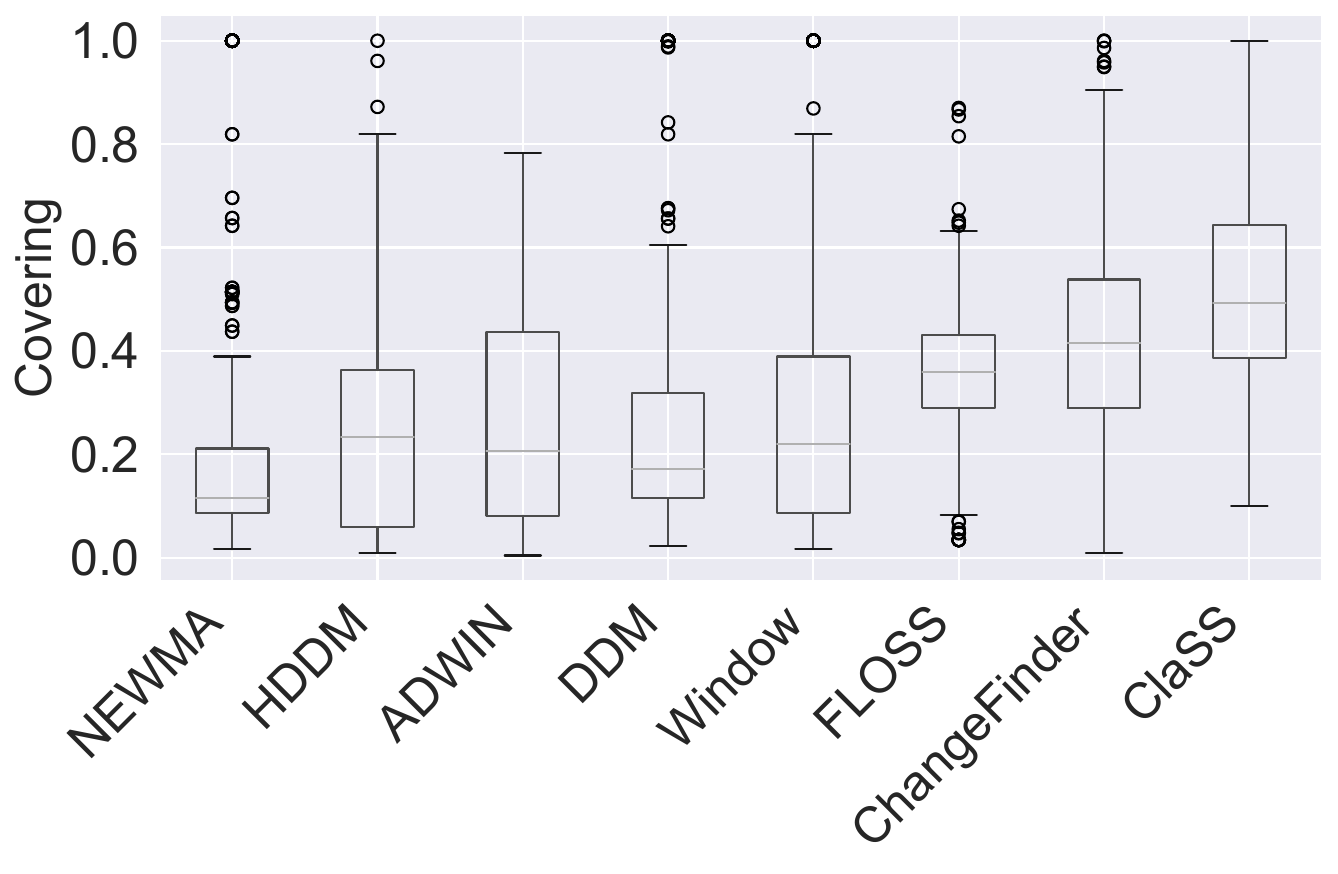}
	\end{minipage}
 
	\caption{Covering segmentation ranks (top) and box plots (bottom) on the $107$ benchmark (left) and $485$ archive (right) TS for ClaSS (lowest rank) and the $8$ competitors.\label{fig:cd_benchmark}
	}
\end{figure}

To aggregate results from multiple data sets into a single ranking, we compute the rank of the score for each method on each TS. We then average the rank of each method across all data sets to obtain its overall rank. Critical Difference (CD) diagrams~\cite{demvsar2006statistical}, such as Figure~\ref{fig:cd_benchmark} (top), are used to statistically assess differences in the mean ranks. The best approaches, which score the lowest average ranks, are shown to the right of the diagram. Approaches that are not significantly different in their ranks are connected by a bar, based on a Nemenyi two-tailed significance test with $\alpha=0.05$.

\paragraph{Competitors}


We compare ClaSS with 8 state-of-the-art competitors (see Table~\ref{tab:competitors}), learning optimal hyper-parameters for all algorithms by testing multiple design choices on $20\%$ randomly chosen benchmark TS (21 out of 107), to prevent overfitting. For a fair comparison, we learned the design choices of ClaSS on the same TS, described in Subsection~\ref{sec:design_choices}. Some of the competitors (Window, BOCD, ChangeFinder and FLOSS) do not specify online segmentation procedures, but only present homogeneity scores for sliding window splits. We learned a threshold for these scores and report splits with a certain quality, considering an exclusion zone (as proposed in~\cite{Gharghabi2018DomainAO}) to prevent series of closely located splits. 

From the CPD literature, we use a discrepancy Window algorithm~\cite{Truong2019SelectiveRO} with a sliding window size of 10 times the annotated subsequence width, a cost function and a threshold as a baseline. We tested autoregressive, Gaussian, kernel, L1, L2, and Mahalanobis cost functions with thresholds from $0.05$ to $0.95$ (steps of $0.05$). Autoregressive cost thresholded at $0.2$ produced the highest mean Covering performance of $51.4\%$ and the most wins among all tested configurations, so we used it as the default hyper-parameter. We include results for the Bayesian method BOCD~\cite{adams2007bayesian}, which is one of the best-performing methods according to~\cite{van2020evaluation}. Testing thresholds for the run length differences in the range of $-50$ to $-500$ (steps of -50) yielded the best results with $-150$, with a mean Covering of $54\%$. We also evaluate ChangeFinder~\cite{Yamanishi2002AUF} and the more recent NEWMA~\cite{Keriven2018NEWMAAN} which use (exponentially weighted) moving averages to detect CPs. ChangeFinder has best-ranking results and a mean Covering performance of $51.3\%$ with a threshold of 50; checking $10$ to $100$ (steps of 10). Regarding NEWMA, we tested quantiles for its adaptive threshold from $0.95$ to $1.0$ (steps of $0.01$), with $1.0$ producing best-scoring results and $46.2\%$ mean Covering.

Additionally, we evaluate DDM~\cite{Gama2004LearningWD}, ADWIN~\cite{Bifet2007LearningFT} and HDDM~\cite{Blanco2015OnlineAN} from drift detection research that represent different approaches, such as monitoring a model's error rate, using adaptive stream statistics or Hoeffding’s inequality, to detect sudden drifts that infer the last completed segment. Both DDM and HDDM have parameters to control the amount of issued drifts, that we set to $20$ (tested $15$ to $30$, steps of 1) and $1e-60$ (checked $1e-10$ to $1e-100$, steps of $1e-10$) with mean Covering performances of $58.4\%$ and $37.5\%$. ADWIN sets a delta value for which we use $0.01$ (tested $0.002$ to $1$, even-valued steps per magnitude) with best-scoring results and a mean Covering of $41.5\%$. We also tried the Page-Hinkley test~\cite{Page1954CONTINUOUSIS}, but could not find a configuration that outputs meaningful results. 

Lastly, we evaluate the data mining approach FLOSS~\cite{gharghabi2017matrix} with a sliding window size of $d=10k$, one of the best-ranking TSS algorithms on the UTSA benchmark~\cite{Gharghabi2018DomainAO,Ermshaus2022ClaSP}. Thresholds for its arc curve were tested from $0.05$ to $0.95$ (steps of $0.05$), with the best-ranking one being $0.45$ at a mean Covering performance of $54.8\%$, and subsequence widths were taken from the annotations.

Note that BOCD processes all so far observed data points of length $n$, ClaSS and FLOSS require a sliding window of size $d$, whereas all other methods either use adaptive / custom-sized sliding windows of length $c \ll d$, or update constant-sized statistics (see Table~\ref{tab:competitors}).

\subsection{Ablation Study} \label{sec:design_choices}

ClaSS has seven major design choices that determine its performance: (a) sliding window size, (b) subsequence width selection method, (c) similarity measure, (d) number of $k$ neighbours used in the streaming $k$-NN, (e) classification score to evaluate the cross-validations, (f) significance level and (g) sample size used for the detection of significant CPs. We tested ClaSS on the same randomly chosen $20\%$ of benchmark TS with varying values (or methods) of each parameter while fixing the others to their default values. We summarize the results of these extensive experiments and refer the interested reader to our supporting website~\cite{ClaSSWebpage} for the raw measurements and visualizations.

\textbf{(a) Sliding Window Size:} We computed ClaSS with sliding window sizes ranging between $1k$ to $20k$ (steps of $1k$) data points. This group of design choices does not show statistically significant differences in ranks, ranging between $76.7\%$ and $81.4\%$ average Covering performances, $13.2\%$ and $17.4\%$ standard deviations and $9$ to $14$ wins. We choose $d=10k$ as a robust default parameter for many scenarios. The user may adapt this parameter, however, to control the throughput of ClaSS (see Subsection~\ref{sec:hyper-param} and~\ref{sec:runtime}). 

\textbf{(b) Window Size Selection:} We tested two whole-series based methods from~\cite{Ermshaus2022WSS}, the most dominant Fourier frequency (FFT) and the highest autocorrelation offset (ACF), as well as two subsequence-based algorithms, Multi-Window-Finder (MWF)~\cite{ImaniMultiWindowFinderDA} and Summary Statistics Subsequence (SuSS)~\cite{Ermshaus2022ClaSP}. Our results show no significant differences between the ranks of the methods. We choose SuSS for WSS in ClaSS, as it achieves the most wins and best mean (standard deviation) Covering performance of $79.1\%$ ($15.6\%$).

\textbf{(c, d) k-Nearest Neighbours:} We evaluated Pearson correlation, Euclidean distance and CID as (dis-)similarity measures and $k \in [1,3,5,7]$. We found no significant differences between rankings; Pearson correlation and a $3$-NN score the best ranks and show the best Covering performance. Therefore, we use both as the default. Users may change the similarity measure to fit specific use cases.

\textbf{(e) Classification Score:} For the sliding window scoring, we assessed the F1 score and accuracy. We used the macro formulation for both scores, which computes them per label and then averages the results, to tackle the inherent class imbalance in the ClaSP calculated by ClaSS. We did not test the ROC/AUC score, used in the batch ClaSP, as it is not computable in constant runtime from the confusion matrix, which is a prerequisite for us to keep the linear runtime complexity in ClaSS. F1 ranks are not significantly better than accuracy, but show best results. Thus, we use it as the default classification score.

\textbf{(f, g) Significance Level:} Lastly, we evaluated significance levels in the range 1e-10 to 1e-100 (with steps of 1e-10) and sample sizes (variable, 10, 100, 1k, 10k) for extracting significant CPs in the sliding window stream segmentation. The variable sample size uses the entire label configuration, as proposed in~\cite{Ermshaus2022ClaSP}. We found that the range of thresholds between 1e-50 and 1e-100 as well as the variable and 1k sample sizes substantially outperformed the other options. The significance level of 1e-50 with 1k sample size achieved the highest mean Covering score and the lowest standard deviation. Thus, we use this configuration in ClaSS.

We conclude that the choice of the sliding window size, the subsequence width, the streaming $k$-NN and the sliding window score in ClaSS have only negligible effects on its performance. For specific domains, users may adjust the significance level to achieve optimal results. This is to be expected, as this parameter directly influences the number of reported CPs, similarly to the thresholds of the competitors.

\begin{table}[t]
 	\caption{Summary Covering performances for ClaSS (best results) and its $8$ competitors on the two benchmarks and six data archives.\label{tab:benchmark_summary}}
	\begin{minipage}{8cm}
	    \captionof*{table}{Benchmarks / Data Archives}
    	\begin{centering}
    		\begin{tabular}{c|ccc}
    			\toprule 			
    			 & mean (in \%) & median (in \%) & std (in \%) \tabularnewline
    			\hline 
    			 ClaSS & $\textbf{81.2}$  / $\textbf{51.5}$ & $\textbf{88.2}$ / $\textbf{49.3}$ & $19.0$ / $17.1$ \tabularnewline
              ChangeFinder & $47.3$ / $42.3$ & $50.0$ / $41.6$ & $23.5$ / $19.7$ \tabularnewline
    			 FLOSS & $52.1$ / $35.6$ & $50.0$ / $35.9$ & $22.7$ / $13.0$ \tabularnewline
              Window & $46.1$ / $29.1$ & $47.4$ / $22.0$ & $24.7$ / $27.7$ \tabularnewline
              DDM & $53.5$ / $26.2$ & $51.3$ / $17.1$ & $16.9$ / $24.5$ \tabularnewline
              BOCD & $48.1$ / - & $49.4$ / - & $19.0$ / - \tabularnewline
              ADWIN & $38.3$ / $26.2$ & $34.2$ / $20.6$ & $20.6$ / $20.5$ \tabularnewline
              HDDM & $36.5$ / $24.6$ & $33.3$ / $23.4$ & $24.8$ / $18.5$ \tabularnewline
              NEWMA & $43.4$ / $21.5$ & $47.4$ / $11.6$ & $20.6$ / $26.2$ \tabularnewline
            \bottomrule 			
    		\end{tabular}
    	\end{centering}
	\end{minipage}
\end{table}

\subsection{Quantitative Analysis} \label{sec:comparative_analysis}

We evaluate the performance of ClaSS and its 8 competitors separately for the two benchmark data sets and six data archives from experimental studies. We remark that the data archives are, by far, the harder scenario as they contain ambiguities, anomalies and signal noise and are up to two orders of magnitude larger than the benchmarks; note that algorithms are not fine-tuned to these conditions. Detailed measurements and visualizations are reported on our supporting website~\cite{ClaSSWebpage}.

\paragraph{Benchmark Data Sets} The CD diagram in Figure~\ref{fig:cd_benchmark} (top left) illustrates the mean Covering ranks. Best results are obtained by ClaSS (1.5) followed by FLOSS (3.5), Window (3.9), DDM (4.0), ChangeFinder (4.0), NEWMA and BOCD (4.1), HDDM (5.0) and ADWIN (5.1). The performance advance of ClaSS is statistically significant, while the differences between the 2nd to 7th-ranking approaches are not. If we consider both benchmarks separately, ClaSS still achieves the best performances, with an insignificant advance for UTSA ($32$ data sets), but a significant advance for TSSB ($75$ data sets).

ClaSS wins or ties in $78$ of the $107$ cases, followed by FLOSS, Window and ChangeFinder each with each $12$ wins/ties, DDM ($10$), NEWMA ($9$), BOCD ($8$), HDDM ($5$) and ADWIN ($4$). Counts do not sum up to $107$ due to ties. ClaSS achieves first place in four subcases of STSS: TS with one segment ($6$ instances), two segments ($46$ instances), at least three segments ($55$ instances), and reoccurring sub-segments ($10$ instances). In a pairwise comparison of ClaSS against every competitor, ClaSS outperforms all competitors in at least $77\%$ of all cases (see~\cite{ClaSSWebpage}).

ClaSS achieves a mean Covering performance of $81.2\%$, with a standard deviation of $19.0\%$. Figure~\ref{fig:cd_benchmark} (bottom left) and Table~\ref{tab:benchmark_summary} show this to be the highest score, with a large margin of $27.7$ pp over the second-best method. The differences in median results are even more pronounced. The summary statistics of the performances are quite stable across the UTSA and TSSB data sets (data shown on~\cite{ClaSSWebpage}). This shows that ClaSS is able to segment TS streams more accurately than its counterparts on the benchmark data sets. 

\begin{figure}[t]
	\begin{minipage}{4cm}
        \includegraphics[width=1.0\columnwidth]{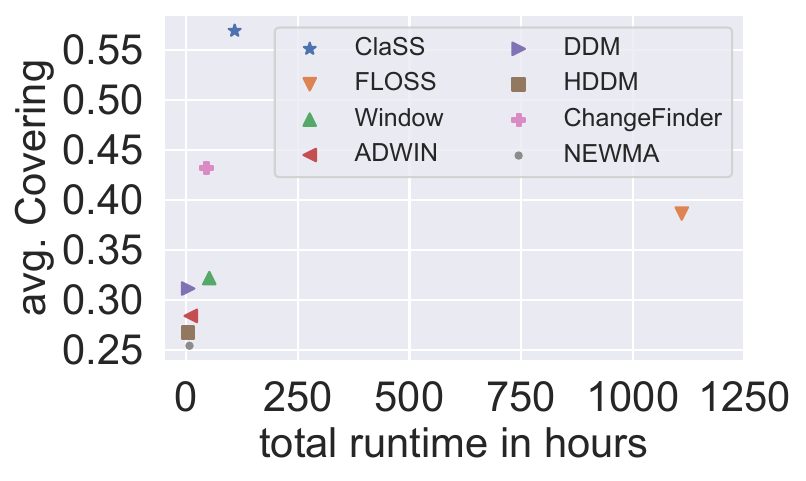}
	\end{minipage}
    \begin{minipage}{4cm}
        \includegraphics[width=1.0\columnwidth]{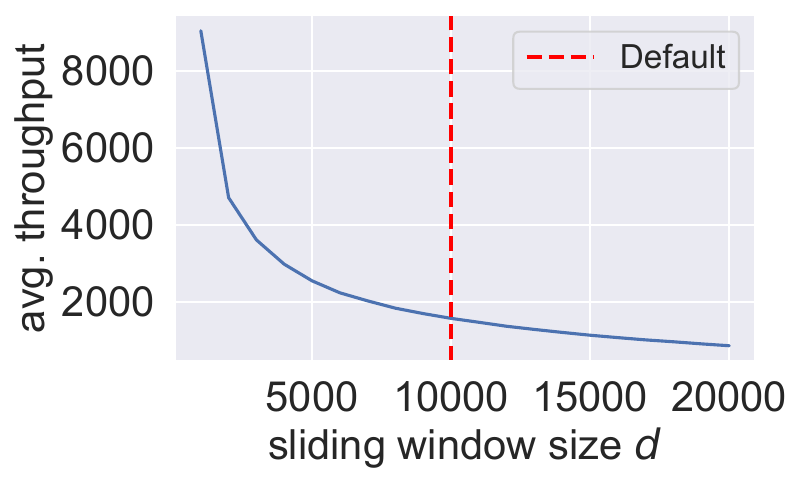}
	\end{minipage}
    \begin{minipage}{4cm}
        \includegraphics[width=1.0\columnwidth]{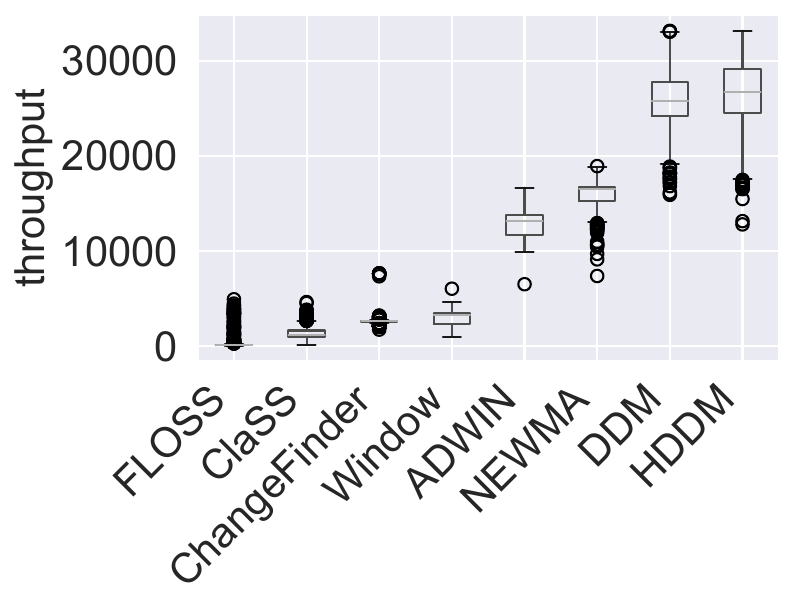}
	\end{minipage}
 	\begin{minipage}{4cm}
        \includegraphics[width=1.0\columnwidth]{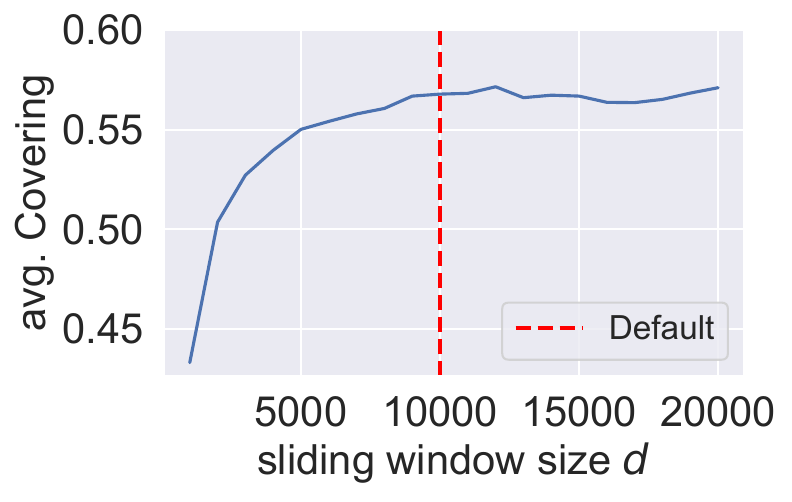}
	\end{minipage}
	\caption{Runtime comparison regarding total time spent vs quality (top left) and standalone data throughput (bottom left) of competitors. Change in throughput (top right) and Covering (bottom right) for different sliding window sizes.\label{fig:scalability}
	}
\end{figure}

\paragraph{Data Archive Sets} For the 485 time series (TS) from the six data archives, ClaSS (2.1) ranks first, followed by ChangeFinder (3.1), FLOSS (3.7), Window (4.5), ADWIN (4.7), DDM (4.8), HDDM (5.0) and NEWMA (5.4) (see Figure~\ref{fig:cd_benchmark} top right). BOCD did not finish within days, and was excluded. Again, ClaSS significantly outperforms its competitors, with the 2nd and 3rd-best competitor ChangeFinder and FLOSS also significantly outperforming the rest. ClaSS ranks first in 5 out of 6 data archives, with 1 significant lead on mHealth and 4 insignificant advances for WESAD, SleepDB, MIT-BIH-VE DB and MIT-BIH-VE-Arr DB. ChangeFinder ranks first on PAMAP, but only with an insignificant difference to ClaSS. We aggregated the average Covering ranks by sensor type and found that ClaSS outperforms its rivals for 7 out of 10 sensors (1 significant); the 3 it performs worse for are electrodermal activity, respiration and body temperature, which are all contained in the WESAD archive and represent just 4 TS per sensor. More annotated TS from these sensors are needed to give a conclusive result on their specific segmentation performance. In a pairwise comparison of ClaSS against the 7 competitors on the data archives, ClaSS achieves the best segmentations in at least $69\%$ instances. 

Considering the summary statistics in Figure~\ref{fig:cd_benchmark} (bottom right) and Table~\ref{tab:benchmark_summary}, all methods drop in mean and median Covering performance but keep similar standard deviations on the data archives compared to the benchmark results. ClaSS scores the highest mean Covering performance of $51.5\%$ and the second-smallest standard deviation of $17.1\%$. The performance improvement of $9.2$ pp compared to the second-best method is substantial, however $18.5$ pp less than for the benchmark results.

\paragraph{Discussion} Our performance analysis shows that ClaSS outperforms 8 other methods in 305 of 592 TS data sets. This superior performance is attributed to two key features of its STSS approach:

\begin{enumerate}
    \item[(a)] ClaSS uses a self-supervised, non-linear $k$-NN classifier for segmenting TS streams, evaluating the likelihood of potential sliding window prefixes being a completed segment. This method is adept at understanding the diverse semantics of signals, unlike the auto-regressive or statistical deviation models used by our competitors, with the exception of FLOSS.
    
    \item[(b)] For identifying CPs, ClaSS utilizes a non-parametric significance test rather than a fixed threshold. This approach allows for more flexible adaptation to different data sets, a strategy not adopted by our competitors except for HDDM.
\end{enumerate}

A real-world example of the impact of these two design choices of ClaSS is explored in Subsection~\ref{sec:usecases}.

\begin{figure}[t]
	\begin{minipage}{4cm}
        \includegraphics[width=1.0\columnwidth]{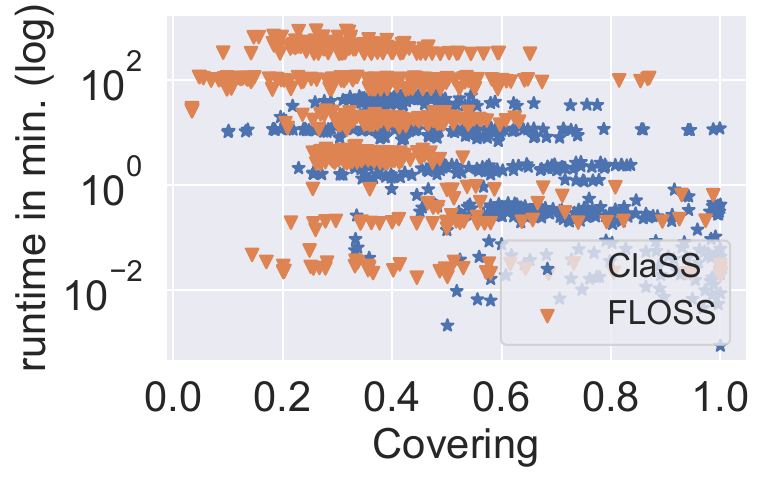}
	\end{minipage}
    \begin{minipage}{4cm}
        \includegraphics[width=1.0\columnwidth]{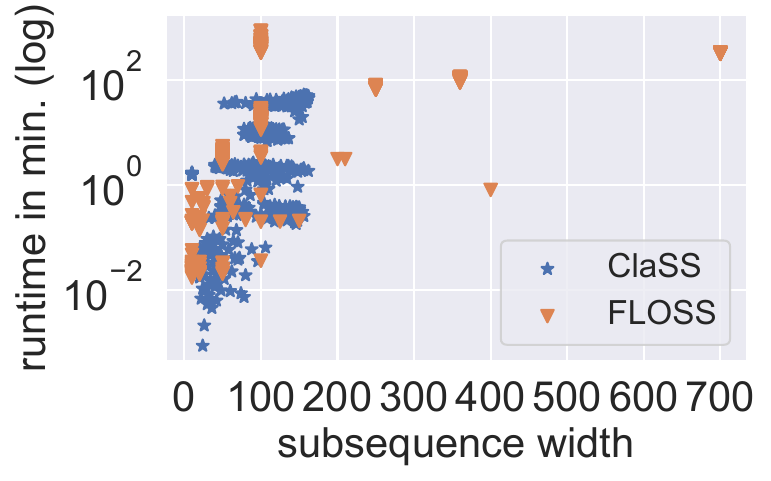}
	\end{minipage}
     \begin{minipage}{4cm}
        \includegraphics[width=1.0\columnwidth]{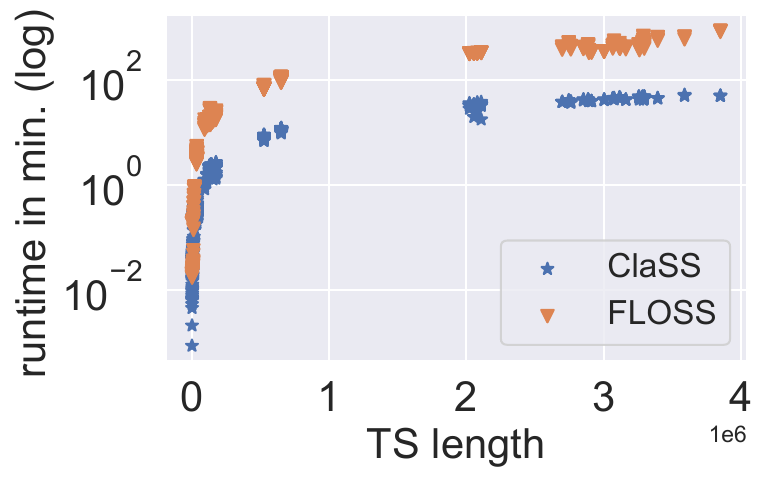}
	\end{minipage}
     \begin{minipage}{4cm}
        \includegraphics[width=1.0\columnwidth]{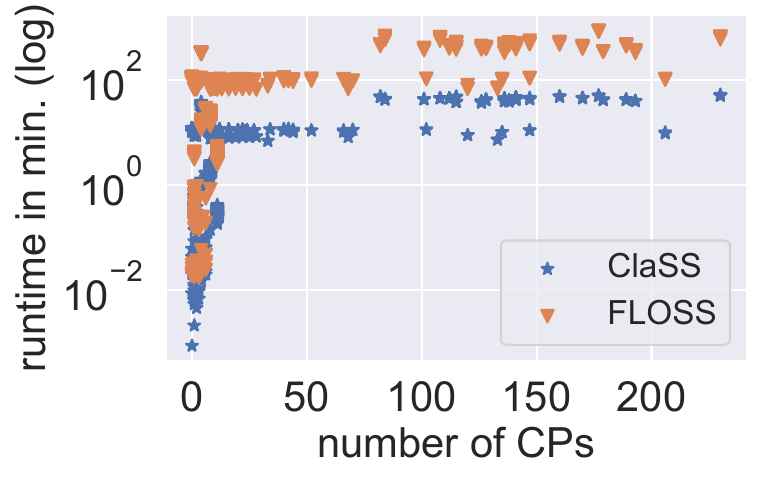}
	\end{minipage}
	\caption{Scalability of ClaSS vs FLOSS considering Covering performance (top left), subsequence width (top right), TS length (bottom left), and number of CPs (bottom right).\label{fig:scalability-comparison}
	}
\end{figure}

\subsection{Runtime and Throughput} \label{sec:runtime}

STSS methods need to process sensor streams in real-time to be useful in practice. We conducted experiments to measure the relationship between runtime and quality as well as data throughput of ClaSS and its competitors on all 592 TS data sets. 

\paragraph{Runtime} As shown in Figure~\ref{fig:scalability} (top left), HDDM is the fastest method (total of 4 hours), followed by DDM (5 hours), NEWMA (7 hours), ADWIN (10 hours), ChangeFinder (45 hours), Window (52 hours), ClaSS (109 hours) and FLOSS (1109 hours), for a total of $3.5$ GB of 64-bit floating-point TS data on a single core. This ranking is roughly aligned with the computational complexities and sliding window sizes of the methods (see Table~\ref{tab:competitors}). The 4 fastest methods build a cluster (bottom left) and produce low average Covering results from $25.4\%$ to $31.1\%$. ChangeFinder and ClaSS trade runtime to score substantially higher average Covering performances of $43.2\%$ and $56.9\%$, while being one order of magnitude slower.  However, ClaSS is more than 10 times faster and $18.3$ pp more accurate than FLOSS, although both methods process the same sliding window. ClaSS needs $36/109$ hours for the bespoke $k$-NN updates. Recomputing dot products increases this runtime to $212$ hours; naive distance calculations take $2513$ hours. Additionally, ClaSS spends $55/109$ hours for the bespoke cross-validations. Using the original implementation from~\cite{Ermshaus2022ClaSP} was stopped after $5755$ hours, segmenting 7 out of 8 data sets. This empirically validates the massive runtime improvements of the central components.

\begin{figure}[t]
    \includegraphics[width=1.0\columnwidth]{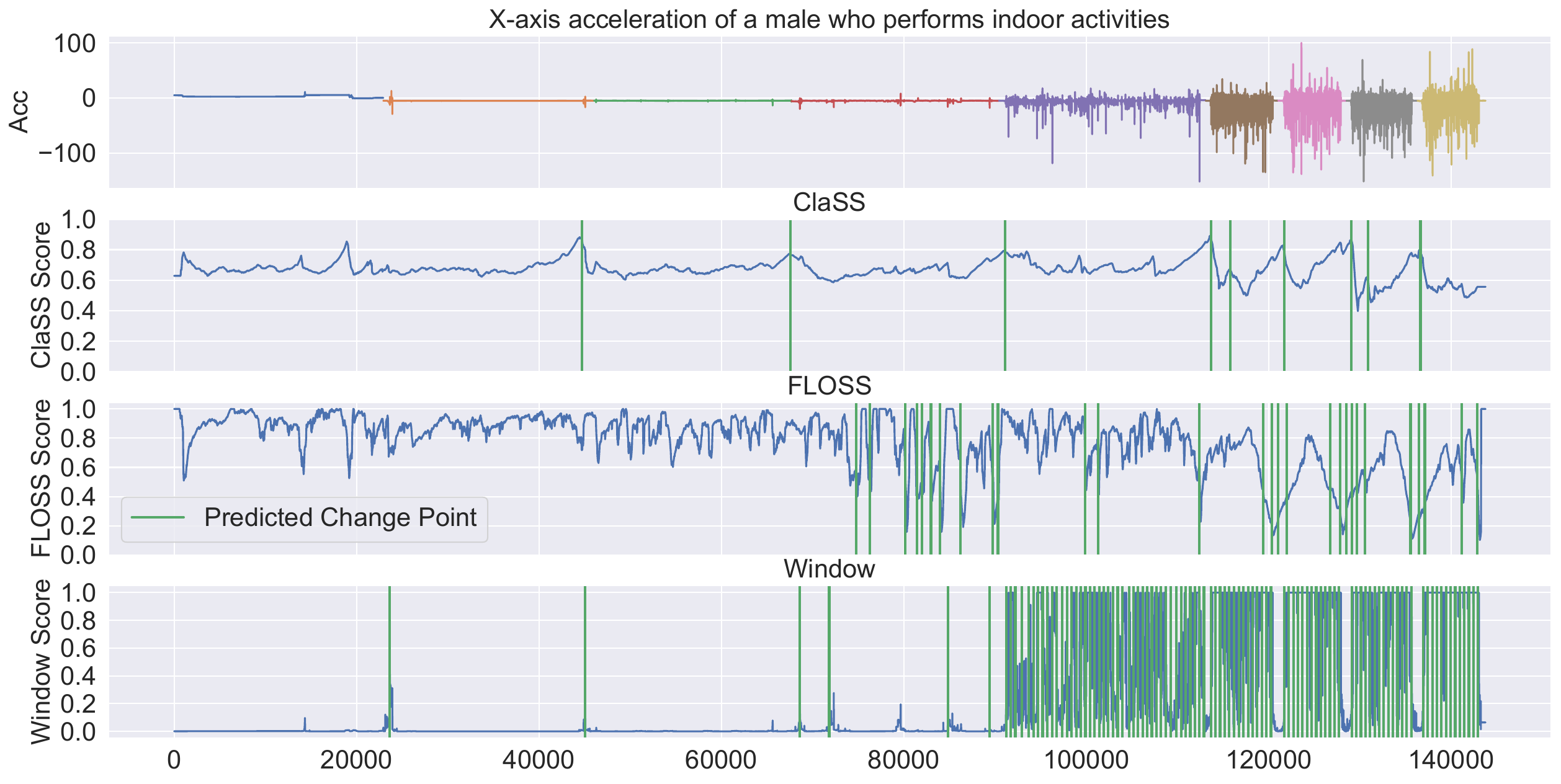}
    \caption{The TS (top) captures the X-axis acceleration of human activity movement~\cite{Reiss2012CreatingAB}. The aggregated profiles for ClaSS, FLOSS and Window (2nd from top to bottom) are illustrated with predicted CPs (green). See~\cite{ClaSSWebpage} for a video.\label{fig:comparative_detection_usecase}
    }
\end{figure}

\paragraph{Standalone Data Throughput} Figure~\ref{fig:scalability} (bottom left) provides a visual representation of the methods' data throughputs when operated in isolation. On average, HDDM and DDM process 26,458 and 26,031 observations per second, followed by NEWMA and ADWIN with 15,949 and 12,958 measurements as well as Window, ChangeFinder, ClaSS and FLOSS with 2,991 down to 378 data points. ClaSS, with an average of 1,408 measurements per second, reaches a maximum of 4,660 observations at times, as its segmentation procedure only scores the unsegmented data points, which leads to throughput peaks. This experiment demonstrates that ClaSS can segment data streams with hundreds of points per second with default parameters using a single core. This is sufficient to segment many IoT or medical sensors that output values in this range. 

\paragraph{Apache Flink Data Throughput} We conducted an evaluation of the ClaSS window operator's data throughput within the Apache Flink framework (see~\cite{ClaSSWebpage}). The execution environment utilized processing time with sequential data processing, as a single instance of a STSS operator can only segment one stream at a time. Each of the $592$ TS was treated as an independent data stream, loaded from RAM, and processed by ClaSS at the maximum possible speed. The output, in turn, was a data stream of CPs. On average, ClaSS processes $1,004$ data points per second (standard deviation: $310$), with peak throughput reaching $2,063$ values. These results are comparable to the standalone setting and demonstrate the integration capability of ClaSS within stream processing engines.

\paragraph{Sliding Window Size} The line plots in Figure~\ref{fig:scalability} (right) illustrate the change in average throughput (top) and Covering (bottom) for sliding window sizes from 1k to 20k (steps of 1k) in ClaSS. The default value of 10k (red line) roughly marks the beginning of converging Covering performance between $56$\% and $57$\%. Cutting it in half to 5k, increases throughput to 2548 data points per second ($1.8$x), loosing $1.8$pp accuracy. Similarly, doubling the default sliding window size to 20k decreases throughput to 863 data points per second ($0.6$x), gaining $0.3$pp of accuracy. We find that sliding window sizes between 5k and 10k retain most of the accuracy while providing scope for throughput optimizations.

\paragraph{Scalability} Figure~\ref{fig:scalability-comparison} shows the scalability of ClaSS vs FLOSS per TS in relation to Covering score, subsequence width, TS length and amount of CPs. We omit a comparison against batch ClaSP, as its quadratic runtime prohibits an application in our experiments (TS with up to 3.9M values). FLOSS and ClaSS share a similar dispersion of runtimes for the variables, shifted by their total difference. For Covering and subsequence width, we do not observe clear relationships. Conversely, for increasing TS length or number of CPs, both algorithms need more runtime. As expected, ClaSS is consistently faster than FLOSS for large TS. On~\cite{ClaSSWebpage}, we additionally show that ClaSS its runtime scales linearly for increasing TS length, which empirically validates its time complexity stated in Subsection~\ref{sec:complexity}. The runtime of ClaSS for segmenting very large offline data archives can probably be accurately predicted using regression.



\begin{figure}[t]
    \includegraphics[width=1.0\columnwidth]{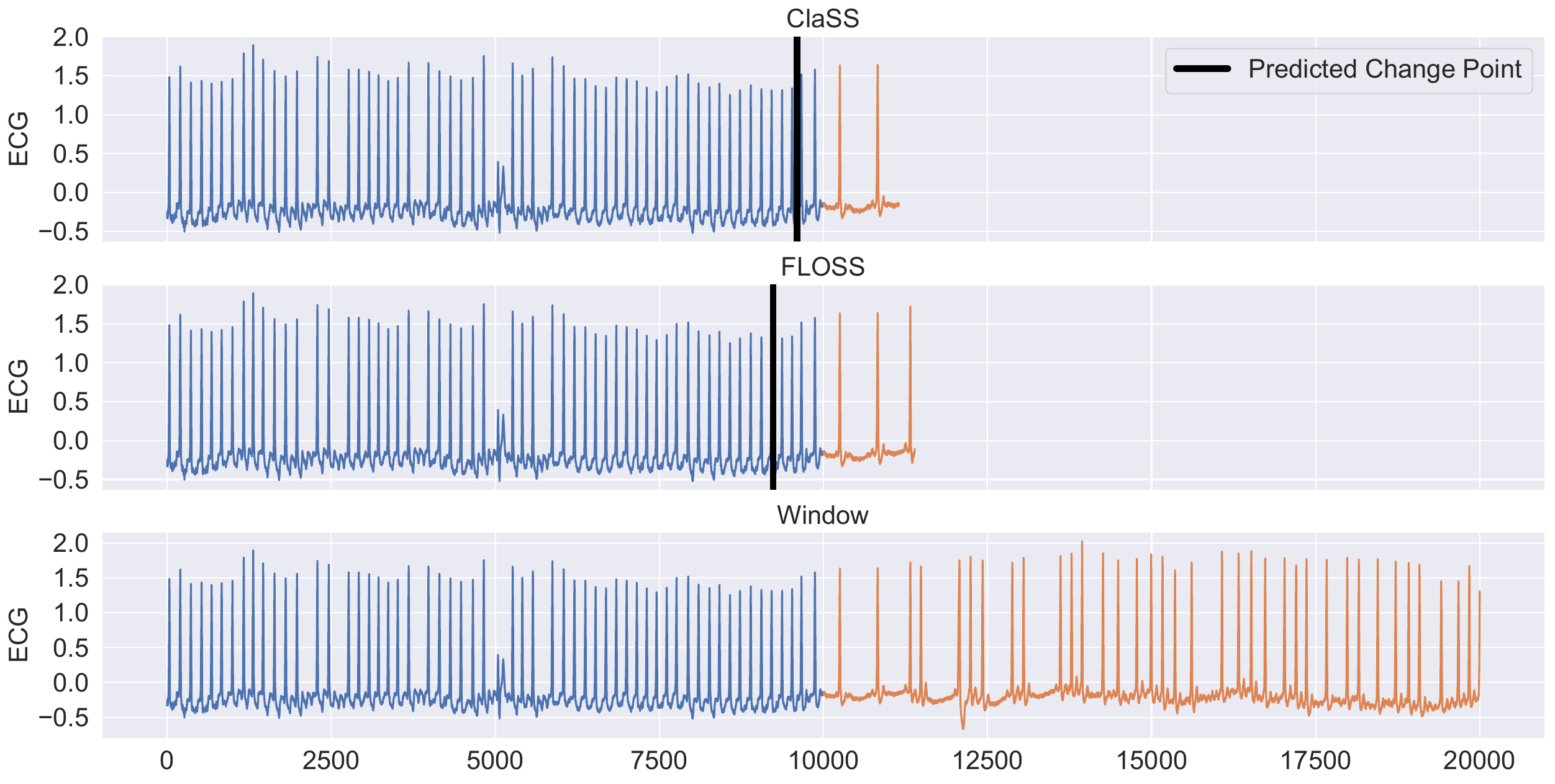}
    \caption{An ECG recording of atrial fibrillations (blue) that transition into premature complexes (orange)~\cite{Moody2001TheIO}. Each of the three TS (top to bottom) illustrate how much data points ClaSS, FLOSS and Window need to ingest to alert the user of the change (black bar).\label{fig:early_detection_usecase}
    }
\end{figure}

\subsection{Use Cases} \label{sec:usecases}

We explore the segmentation results of two interesting use cases to show the characteristics of ClaSS compared to FLOSS and Window (2nd and 3rd best benchmark competitors). Both are examples from two of the six data archives.

\subsubsection*{Human Activity Recognition}

Human activity recognition (HAR) is an important subfield of ubiquitous sensing, with applications in medical condition monitoring and decision-support in tactical scenarios~\cite{Lara2013ASO}. Figure~\ref{fig:comparative_detection_usecase} illustrates an example of HAR, showing accelerometer readings from a 26-year-old male, in the PAMAP data archive~\cite{Reiss2012CreatingAB}, performing a sequence of 9 activities (top). We computed the profiles and predicted CPs using ClaSS, FLOSS and Window (2nd from top to bottom) and visualized the max-aggregated scores for ClaSS and Window and the min-aggregated values for FLOSS. A video of ClaSS' real-time segmentation is available online~\cite{ClaSSWebpage}. ClaSS has a smooth score profile with accurate predictions, missing only a very subtle change from lying to sitting and having two false positives. FLOSS generated a noisy arc curve with more false positives, related to its greedy CP extraction algorithm. Window accurately detected the first four activity transitions, but then had many false positives due to the discrepancy measure misjudging the signal. This use case highlights the accuracy and adaptability of ClaSS, as well as its interpretability for human inspection.

\subsubsection*{Early Streaming Time Series Segmentation}

Extracting the last completed segment as soon as possible is a new and mostly unstudied problem that evaluates methods based on the accuracy of CPs and the time required to report them to the user. This is particularly relevant in domains where precision and fast response times are necessary, such as seismography and cardiology. Figure~\ref{fig:early_detection_usecase} illustrates the ECG signal of a 68-year-old male patient from the MIT-BIH-Arr DB~\cite{Moody2001TheIO} undergoing an onset of atrial fibrillations that changes into premature complexes. The ingested data points for ClaSS, FLOSS, and Window, as well as their predictions (black bars) are plotted to show the amount of observations needed to alert the user. ClaSS accurately predicts the changing conditions after only two heart beats, while FLOSS needs three beats to provide a good prediction. Window, however, completely misses the change. This data set demonstrates that ClaSS is not only accurate, but also requires little time to report the detected CPs. In future research, a benchmark study should be conducted to quantitatively evaluate early segmentation.  

\section{Related Work}

In the last two decades, a wealth of benchmarks~\cite{Paparrizos2022TSBUADAE}, databases~\cite{Pelkonen2015GorillaAF}, indices~\cite{Echihabi2022HerculesAD}, compression algorithms~\cite{Liakos2022ChimpEL}, and data analytics~\cite{Pan2022NLCSC} have been developed by TS management and mining research. This research is driven by the rapidly growing amount of sensor data from IoT devices in \emph{smart} applications for environments, healthcare or factories~\cite{Krishnamurthi2020AnOO}. Sensors, found in wearables or fixed installations, include e.g. accelerometers, thermometers, or optical sensors. Their data, sampled at varying rates, is wirelessly transmitted via Wi-Fi, Bluetooth, or NFC to edge analytics for initial pre-processing and fusion, before being sent to the cloud for advanced analysis~\cite{Santos2013DiAlDS}.

STSS is a complex preprocessing step in many IoT workflows and has been extensively researched in different settings, e.g. on edge devices~\cite{Dbski2021AdaptiveSO}, for smart homes~\cite{Wan2015DynamicSE}, and as a part in integrated HAR systems~\cite{Laguna2011ADS,Cho2015AutomaticSD,Triboan2017SemanticSO}. Such workflows are typically implemented with streaming platforms such as Apache Flink, Spark, or Storm to manage and process vast amounts of TS data in real-time. Stream processing systems mainly differ regarding their processing models, such as one-at-a-time and micro-batch, issued value delivery guarantees (at-least vs. exactly-once) and order~\cite{Wingerath2016RealtimeSP}. Karimov et al.~\cite{Karimov2019BenchmarkingDS} conducted a benchmark comparing these systems in terms of data skew, data arrival fluctuations, latency, and throughput. They found that no single platform consistently outperformed the others, with each possessing unique advantages and disadvantages. For instance, Flink exhibits the lowest average latency but is less effective at handling skewed data compared to Spark. Gehring et al.~\cite{Gehring2019ACO} explored qualitative criteria, such as functionality, simplicity, and documentation, when developing TS analytics using Flink and Spark. Their findings indicate that Flink's development API, evaluation, and visualization functionality are better suited for TS analysis workflows. Consequently, we implemented ClaSS in Flink for integration with its stream processing system.

Besides its practical application, STSS has been studied conceptually as CP and drift detection problems~\cite{Kifer2004DetectingCI,Gama2014ASO}. These formalizations focus on the point at which one segment changes into another. Algorithms monitor the shape or value distributions of sliding windows from a TS stream and report CPs once they substantially differ. The literature differentiates multiple categories of methods~\cite{Aminikhanghahi2016ASO,Truong2019SelectiveRO}. Parametric techniques measure the change in a signal's assumed probability distribution. Implementations include Window~\cite{Truong2019SelectiveRO}, DDM~\cite{Gama2004LearningWD}, ADWIN~\cite{Bifet2007LearningFT} or BOCD~\cite{adams2007bayesian}, which estimates the posterior distribution for the observations since the last CP and can be extended to accommodate for short, gradual changes~\cite{Draayer2021ReevaluatingTC}. Non-parametric approaches do not assume a specific model, and instead compute kernels, distances or rankings on the stream to quantify drift. Algorithms include FLOSS~\cite{Gharghabi2018DomainAO}, which estimates the density of homogenous regions using nearest-neighbour arcs, or our proposed method ClaSS, which utilizes self-supervised learning~\cite{Hido2008UnsupervisedCA}, and makes few assumptions about segments, e.g. being mutually dissimilar. This is in contrast to the aforementioned existing methods, which are either domain-specific, parametric or lack a robust segmentation procedure to handle fluctuations in observations.

\section{Conclusion} \label{sec:conclusion}

We proposed ClaSS, a novel algorithm with minimal assumptions for streaming time series segmentation (STSS) that is amenable to human inspection. Our extensive experiments demonstrate that it sets the new state of the art on two benchmarks with 107 TS, five out of six data archives from experimental studies with 485 TS, and is fast and scalable. In addition to the streaming setting, ClaSS can also be used for very long TS in the batch scenario where computationally expensive TSS algorithms become infeasible. 

Besides its strengths, weaknesses include the initial time points needed to determine the subsequence width and the dependence on the predictive power of the $k$-NN classifier. ClaSS is solely applicable to univariate TS. Many real-world use cases, however, capture processes with a multitude of sensors, where temporal patterns are distributed across various channels. It also has less throughput compared to some competitors, which restricts its applicability for sensors with extremely high sampling rates.

In future work, we plan to extend ClaSS to the multivariate setting, exploring sensor fusion and dimension selection to improve accuracy. We also plan to to accelerate the streaming $k$-NN calculation and significance test by simultaneously  working on sliding window partitions using multi-threading or GPUs.

\bibliographystyle{ACM-Reference-Format}
\bibliography{main}


\begin{thebibliography}{71}


\ifx \showCODEN    \undefined \def \showCODEN     #1{\unskip}     \fi
\ifx \showDOI      \undefined \def \showDOI       #1{#1}\fi
\ifx \showISBNx    \undefined \def \showISBNx     #1{\unskip}     \fi
\ifx \showISBNxiii \undefined \def \showISBNxiii  #1{\unskip}     \fi
\ifx \showISSN     \undefined \def \showISSN      #1{\unskip}     \fi
\ifx \showLCCN     \undefined \def \showLCCN      #1{\unskip}     \fi
\ifx \shownote     \undefined \def \shownote      #1{#1}          \fi
\ifx \showarticletitle \undefined \def \showarticletitle #1{#1}   \fi
\ifx \showURL      \undefined \def \showURL       {\relax}        \fi
\providecommand\bibfield[2]{#2}
\providecommand\bibinfo[2]{#2}
\providecommand\natexlab[1]{#1}
\providecommand\showeprint[2][]{arXiv:#2}

\bibitem[\protect\citeauthoryear{Adams, Alonso, Atkin, Banning, Bhola, Buskens,
  Chen, Chen, Chung, Jia, Sakharov, Talbot, Taylor, and Tart}{Adams
  et~al\mbox{.}}{2020}]%
        {Adams2020MonarchGP}
\bibfield{author}{\bibinfo{person}{Colin Adams}, \bibinfo{person}{Luis Alonso},
  \bibinfo{person}{Benjamin Atkin}, \bibinfo{person}{John~P. Banning},
  \bibinfo{person}{Sumeer Bhola}, \bibinfo{person}{Richard~W. Buskens},
  \bibinfo{person}{Ming Chen}, \bibinfo{person}{Xi Chen}, \bibinfo{person}{Yoo
  Chung}, \bibinfo{person}{Qin Jia}, \bibinfo{person}{Nick Sakharov},
  \bibinfo{person}{George Talbot}, \bibinfo{person}{Nick Taylor}, {and}
  \bibinfo{person}{Adam Tart}.} \bibinfo{year}{2020}\natexlab{}.
\newblock \showarticletitle{Monarch: Google's Planet-Scale In-Memory Time
  Series Database}.
\newblock \bibinfo{journal}{\emph{Proc. VLDB Endow.}}  \bibinfo{volume}{13}
  (\bibinfo{year}{2020}), \bibinfo{pages}{3181--3194}.
\newblock


\bibitem[\protect\citeauthoryear{Adams and MacKay}{Adams and MacKay}{2007}]%
        {adams2007bayesian}
\bibfield{author}{\bibinfo{person}{Ryan~Prescott Adams} {and}
  \bibinfo{person}{David~JC MacKay}.} \bibinfo{year}{2007}\natexlab{}.
\newblock \showarticletitle{Bayesian online changepoint detection}.
\newblock \bibinfo{journal}{\emph{arXiv preprint arXiv:0710.3742}}
  (\bibinfo{year}{2007}).
\newblock


\bibitem[\protect\citeauthoryear{Aminikhanghahi and Cook}{Aminikhanghahi and
  Cook}{2016}]%
        {Aminikhanghahi2016ASO}
\bibfield{author}{\bibinfo{person}{Samaneh Aminikhanghahi} {and}
  \bibinfo{person}{Diane~Joyce Cook}.} \bibinfo{year}{2016}\natexlab{}.
\newblock \showarticletitle{A survey of methods for time series change point
  detection}.
\newblock \bibinfo{journal}{\emph{Knowledge and Information Systems}}
  \bibinfo{volume}{51} (\bibinfo{year}{2016}), \bibinfo{pages}{339--367}.
\newblock


\bibitem[\protect\citeauthoryear{Ba{\~n}os, Garc{\'i}a, Terriza, Damas,
  Pomares, Rojas, Saez, and Villalonga}{Ba{\~n}os et~al\mbox{.}}{2014}]%
        {Baos2014mHealthDroidAN}
\bibfield{author}{\bibinfo{person}{Oresti Ba{\~n}os}, \bibinfo{person}{Rafael
  Garc{\'i}a}, \bibinfo{person}{Juan Antonio~Holgado Terriza},
  \bibinfo{person}{Miguel Damas}, \bibinfo{person}{H{\'e}ctor Pomares},
  \bibinfo{person}{Ignacio Rojas}, \bibinfo{person}{Alejandro Saez}, {and}
  \bibinfo{person}{Claudia Villalonga}.} \bibinfo{year}{2014}\natexlab{}.
\newblock \showarticletitle{mHealthDroid: A Novel Framework for Agile
  Development of Mobile Health Applications}. In
  \bibinfo{booktitle}{\emph{IWAAL}}.
\newblock


\bibitem[\protect\citeauthoryear{Ba{\~n}os, Villalonga, Garc{\'i}a, Saez,
  Damas, Holgado-Terriza, Lee, Pomares, and Rojas}{Ba{\~n}os
  et~al\mbox{.}}{2015}]%
        {Baos2015DesignIA}
\bibfield{author}{\bibinfo{person}{Oresti Ba{\~n}os}, \bibinfo{person}{Claudia
  Villalonga}, \bibinfo{person}{Rafael Garc{\'i}a}, \bibinfo{person}{Alejandro
  Saez}, \bibinfo{person}{Miguel Damas}, \bibinfo{person}{Juan~Antonio
  Holgado-Terriza}, \bibinfo{person}{Sungyong Lee}, \bibinfo{person}{H{\'e}ctor
  Pomares}, {and} \bibinfo{person}{Ignacio Rojas}.}
  \bibinfo{year}{2015}\natexlab{}.
\newblock \showarticletitle{Design, implementation and validation of a novel
  open framework for agile development of mobile health applications}.
\newblock \bibinfo{journal}{\emph{BioMedical Engineering OnLine}}
  \bibinfo{volume}{14} (\bibinfo{year}{2015}), \bibinfo{pages}{S6 -- S6}.
\newblock


\bibitem[\protect\citeauthoryear{Batista, Keogh, Tataw, and Souza}{Batista
  et~al\mbox{.}}{2013}]%
        {Batista2013CIDAE}
\bibfield{author}{\bibinfo{person}{Gustavo E. A. P.~A. Batista},
  \bibinfo{person}{Eamonn~J. Keogh}, \bibinfo{person}{Oben~M. Tataw}, {and}
  \bibinfo{person}{Vinicius M.~A. Souza}.} \bibinfo{year}{2013}\natexlab{}.
\newblock \showarticletitle{CID: an efficient complexity-invariant distance for
  time series}.
\newblock \bibinfo{journal}{\emph{Data Mining and Knowledge Discovery}}
  \bibinfo{volume}{28} (\bibinfo{year}{2013}), \bibinfo{pages}{634--669}.
\newblock


\bibitem[\protect\citeauthoryear{Beyreuther, Barsch, Krischer, Megies, Behr,
  and Wassermann}{Beyreuther et~al\mbox{.}}{2010}]%
        {Beyreuther2010ObsPyAP}
\bibfield{author}{\bibinfo{person}{M. Beyreuther}, \bibinfo{person}{Robert
  Barsch}, \bibinfo{person}{Lion Krischer}, \bibinfo{person}{Tobias Megies},
  \bibinfo{person}{Yannik Behr}, {and} \bibinfo{person}{Joachim Wassermann}.}
  \bibinfo{year}{2010}\natexlab{}.
\newblock \showarticletitle{ObsPy: A Python Toolbox for Seismology}.
\newblock \bibinfo{journal}{\emph{Seismological Research Letters}}
  \bibinfo{volume}{81} (\bibinfo{year}{2010}), \bibinfo{pages}{530--533}.
\newblock


\bibitem[\protect\citeauthoryear{Bifet and Gavald{\`a}}{Bifet and
  Gavald{\`a}}{2007}]%
        {Bifet2007LearningFT}
\bibfield{author}{\bibinfo{person}{Albert Bifet} {and} \bibinfo{person}{Ricard
  Gavald{\`a}}.} \bibinfo{year}{2007}\natexlab{}.
\newblock \showarticletitle{Learning from Time-Changing Data with Adaptive
  Windowing}. In \bibinfo{booktitle}{\emph{SDM}}.
\newblock


\bibitem[\protect\citeauthoryear{Blanco, del Campo-{\'A}vila,
  Ramos-Jim{\'e}nez, Bueno, D{\'i}az, and Mota}{Blanco et~al\mbox{.}}{2015}]%
        {Blanco2015OnlineAN}
\bibfield{author}{\bibinfo{person}{Isvani Inocencio~Fr{\'i}as Blanco},
  \bibinfo{person}{Jos{\'e} del Campo-{\'A}vila}, \bibinfo{person}{Gonzalo
  Ramos-Jim{\'e}nez}, \bibinfo{person}{Rafael~Morales Bueno},
  \bibinfo{person}{Agust{\'i}n Alejandro~Ortiz D{\'i}az}, {and}
  \bibinfo{person}{Yail{\'e}~Caballero Mota}.} \bibinfo{year}{2015}\natexlab{}.
\newblock \showarticletitle{Online and Non-Parametric Drift Detection Methods
  Based on Hoeffding’s Bounds}.
\newblock \bibinfo{journal}{\emph{IEEE Transactions on Knowledge and Data
  Engineering}}  \bibinfo{volume}{27} (\bibinfo{year}{2015}),
  \bibinfo{pages}{810--823}.
\newblock


\bibitem[\protect\citeauthoryear{Cho, An, Hong, and Lee}{Cho
  et~al\mbox{.}}{2015}]%
        {Cho2015AutomaticSD}
\bibfield{author}{\bibinfo{person}{Hyunjeong Cho}, \bibinfo{person}{Jihoon An},
  \bibinfo{person}{Intaek Hong}, {and} \bibinfo{person}{Younghee Lee}.}
  \bibinfo{year}{2015}\natexlab{}.
\newblock \showarticletitle{Automatic Sensor Data Stream Segmentation for
  Real-time Activity Prediction in Smart Spaces}.
\newblock \bibinfo{journal}{\emph{Proceedings of the 2015 Workshop on IoT
  challenges in Mobile and Industrial Systems}} (\bibinfo{year}{2015}).
\newblock


\bibitem[\protect\citeauthoryear{{ClaSS Code and Raw Results}}{{ClaSS Code and
  Raw Results}}{2023}]%
        {ClaSSWebpage}
\bibfield{author}{\bibinfo{person}{{ClaSS Code and Raw Results}}.}
  \bibinfo{year}{2023}\natexlab{}.
\newblock
  \bibinfo{howpublished}{\url{https://github.com/ermshaua/classification-score-stream}}.
\newblock


\bibitem[\protect\citeauthoryear{Dau, Bagnall, Kamgar, Yeh, Zhu, Gharghabi,
  Ratanamahatana, and Keogh}{Dau et~al\mbox{.}}{2019}]%
        {UCRClassification}
\bibfield{author}{\bibinfo{person}{Hoang~Anh Dau}, \bibinfo{person}{Anthony~J.
  Bagnall}, \bibinfo{person}{Kaveh Kamgar}, \bibinfo{person}{Chin-Chia~Michael
  Yeh}, \bibinfo{person}{Yan Zhu}, \bibinfo{person}{Shaghayegh Gharghabi},
  \bibinfo{person}{Chotirat Ratanamahatana}, {and} \bibinfo{person}{Eamonn~J.
  Keogh}.} \bibinfo{year}{2019}\natexlab{}.
\newblock \showarticletitle{The UCR time series archive}.
\newblock \bibinfo{journal}{\emph{IEEE/CAA Journal of Automatica Sinica}}
  \bibinfo{volume}{6} (\bibinfo{year}{2019}), \bibinfo{pages}{1293--1305}.
\newblock


\bibitem[\protect\citeauthoryear{D{\k{e}}bski and Dre{\.z}ewski}{D{\k{e}}bski
  and Dre{\.z}ewski}{2021}]%
        {Dbski2021AdaptiveSO}
\bibfield{author}{\bibinfo{person}{Roman D{\k{e}}bski} {and}
  \bibinfo{person}{Rafa{\l} Dre{\.z}ewski}.} \bibinfo{year}{2021}\natexlab{}.
\newblock \showarticletitle{Adaptive Segmentation of Streaming Sensor Data on
  Edge Devices}.
\newblock \bibinfo{journal}{\emph{Sensors (Basel, Switzerland)}}
  \bibinfo{volume}{21} (\bibinfo{year}{2021}).
\newblock


\bibitem[\protect\citeauthoryear{Dem{\v{s}}ar}{Dem{\v{s}}ar}{2006}]%
        {demvsar2006statistical}
\bibfield{author}{\bibinfo{person}{Janez Dem{\v{s}}ar}.}
  \bibinfo{year}{2006}\natexlab{}.
\newblock \showarticletitle{{Statistical Comparisons of Classifiers over
  Multiple Data Sets}}.
\newblock \bibinfo{journal}{\emph{{The Journal of Machine Learning Research}}}
  \bibinfo{volume}{7} (\bibinfo{year}{2006}), \bibinfo{pages}{1--30}.
\newblock


\bibitem[\protect\citeauthoryear{Draayer, Cao, and Hao}{Draayer
  et~al\mbox{.}}{2021}]%
        {Draayer2021ReevaluatingTC}
\bibfield{author}{\bibinfo{person}{Erick Draayer}, \bibinfo{person}{Huiping
  Cao}, {and} \bibinfo{person}{Yifan Hao}.} \bibinfo{year}{2021}\natexlab{}.
\newblock \showarticletitle{Reevaluating the Change Point Detection Problem
  with Segment-based Bayesian Online Detection}.
\newblock \bibinfo{journal}{\emph{Proceedings of the 30th ACM International
  Conference on Information \& Knowledge Management}} (\bibinfo{year}{2021}).
\newblock


\bibitem[\protect\citeauthoryear{Echihabi, Fatourou, Zoumpatianos, Palpanas,
  and Benbrahim}{Echihabi et~al\mbox{.}}{2022}]%
        {Echihabi2022HerculesAD}
\bibfield{author}{\bibinfo{person}{Karima Echihabi}, \bibinfo{person}{Panagiota
  Fatourou}, \bibinfo{person}{Kostas Zoumpatianos}, \bibinfo{person}{Themis
  Palpanas}, {and} \bibinfo{person}{Houda Benbrahim}.}
  \bibinfo{year}{2022}\natexlab{}.
\newblock \showarticletitle{Hercules Against Data Series Similarity Search}.
\newblock \bibinfo{journal}{\emph{Proc. VLDB Endow.}}  \bibinfo{volume}{15}
  (\bibinfo{year}{2022}), \bibinfo{pages}{2005--2018}.
\newblock


\bibitem[\protect\citeauthoryear{Ermshaus, Sch{\"a}fer, and Leser}{Ermshaus
  et~al\mbox{.}}{2022}]%
        {Ermshaus2022WSS}
\bibfield{author}{\bibinfo{person}{Arik Ermshaus}, \bibinfo{person}{Patrick
  Sch{\"a}fer}, {and} \bibinfo{person}{Ulf Leser}.}
  \bibinfo{year}{2022}\natexlab{}.
\newblock \showarticletitle{Window Size Selection In Unsupervised Time Series
  Analytics: A Review and Benchmark}.
\newblock \bibinfo{journal}{\emph{7th Workshop on Advanced Analytics and
  Learning on Temporal Data}} (\bibinfo{year}{2022}).
\newblock


\bibitem[\protect\citeauthoryear{Ermshaus, Sch{\"a}fer, and Leser}{Ermshaus
  et~al\mbox{.}}{2023}]%
        {Ermshaus2022ClaSP}
\bibfield{author}{\bibinfo{person}{Arik Ermshaus}, \bibinfo{person}{Patrick
  Sch{\"a}fer}, {and} \bibinfo{person}{Ulf Leser}.}
  \bibinfo{year}{2023}\natexlab{}.
\newblock \showarticletitle{ClaSP: parameter-free time series segmentation}.
\newblock \bibinfo{journal}{\emph{Data Mining and Knowledge Discovery}}
  \bibinfo{volume}{37} (\bibinfo{year}{2023}), \bibinfo{pages}{1262 -- 1300}.
\newblock


\bibitem[\protect\citeauthoryear{Gama, Medas, Castillo, and Rodrigues}{Gama
  et~al\mbox{.}}{2004}]%
        {Gama2004LearningWD}
\bibfield{author}{\bibinfo{person}{Jo{\~a}o Gama}, \bibinfo{person}{Pedro
  Medas}, \bibinfo{person}{Gladys Castillo}, {and}
  \bibinfo{person}{Pedro~Pereira Rodrigues}.} \bibinfo{year}{2004}\natexlab{}.
\newblock \showarticletitle{Learning with Drift Detection}. In
  \bibinfo{booktitle}{\emph{Brazilian Symposium on Artificial Intelligence}}.
\newblock


\bibitem[\protect\citeauthoryear{Gama and Rodrigues}{Gama and
  Rodrigues}{2007}]%
        {Gama2007data}
\bibfield{author}{\bibinfo{person}{Jo{\~a}o Gama} {and}
  \bibinfo{person}{Pedro~Pereira Rodrigues}.} \bibinfo{year}{2007}\natexlab{}.
\newblock \showarticletitle{Data stream processing}.
\newblock In \bibinfo{booktitle}{\emph{Learning from Data Streams}}.
  \bibinfo{publisher}{Springer}, \bibinfo{pages}{25--39}.
\newblock


\bibitem[\protect\citeauthoryear{Gama, Žliobaitė, Bifet, Pechenizkiy, and
  Bouchachia}{Gama et~al\mbox{.}}{2014}]%
        {Gama2014ASO}
\bibfield{author}{\bibinfo{person}{Jo{\~a}o Gama}, \bibinfo{person}{Indrė
  Žliobaitė}, \bibinfo{person}{Albert Bifet}, \bibinfo{person}{Mykola
  Pechenizkiy}, {and} \bibinfo{person}{A. Bouchachia}.}
  \bibinfo{year}{2014}\natexlab{}.
\newblock \showarticletitle{A survey on concept drift adaptation}.
\newblock \bibinfo{journal}{\emph{ACM Computing Surveys (CSUR)}}
  \bibinfo{volume}{46} (\bibinfo{year}{2014}), \bibinfo{pages}{1 -- 37}.
\newblock


\bibitem[\protect\citeauthoryear{Gehring, Charfuelan, and Markl}{Gehring
  et~al\mbox{.}}{2019}]%
        {Gehring2019ACO}
\bibfield{author}{\bibinfo{person}{Melissa Gehring}, \bibinfo{person}{Marcela
  Charfuelan}, {and} \bibinfo{person}{Volker Markl}.}
  \bibinfo{year}{2019}\natexlab{}.
\newblock \showarticletitle{A Comparison of Distributed Stream Processing
  Systems for Time Series Analysis}. In
  \bibinfo{booktitle}{\emph{Datenbanksysteme f{\"u}r Business, Technologie und
  Web}}.
\newblock


\bibitem[\protect\citeauthoryear{Gharghabi, Ding, Yeh, Kamgar, Ulanova, and
  Keogh}{Gharghabi et~al\mbox{.}}{2017}]%
        {gharghabi2017matrix}
\bibfield{author}{\bibinfo{person}{Shaghayegh Gharghabi},
  \bibinfo{person}{Yifei Ding}, \bibinfo{person}{Chin-Chia~Michael Yeh},
  \bibinfo{person}{Kaveh Kamgar}, \bibinfo{person}{Liudmila Ulanova}, {and}
  \bibinfo{person}{Eamonn Keogh}.} \bibinfo{year}{2017}\natexlab{}.
\newblock \showarticletitle{Matrix profile VIII: domain agnostic online
  semantic segmentation at superhuman performance levels}. In
  \bibinfo{booktitle}{\emph{ICDM}}. IEEE, \bibinfo{pages}{117--126}.
\newblock


\bibitem[\protect\citeauthoryear{Gharghabi, Yeh, Ding, Ding, Hibbing, LaMunion,
  Kaplan, Crouter, and Keogh}{Gharghabi et~al\mbox{.}}{2018}]%
        {Gharghabi2018DomainAO}
\bibfield{author}{\bibinfo{person}{Shaghayegh Gharghabi},
  \bibinfo{person}{Chin-Chia~Michael Yeh}, \bibinfo{person}{Yifei Ding},
  \bibinfo{person}{Wei Ding}, \bibinfo{person}{Paul~R. Hibbing},
  \bibinfo{person}{Samuel~R LaMunion}, \bibinfo{person}{Andrew Kaplan},
  \bibinfo{person}{Scott~E. Crouter}, {and} \bibinfo{person}{Eamonn~J. Keogh}.}
  \bibinfo{year}{2018}\natexlab{}.
\newblock \showarticletitle{Domain agnostic online semantic segmentation for
  multi-dimensional time series}.
\newblock \bibinfo{journal}{\emph{Data Mining and Knowledge Discovery}}
  \bibinfo{volume}{33} (\bibinfo{year}{2018}), \bibinfo{pages}{96 -- 130}.
\newblock


\bibitem[\protect\citeauthoryear{Goldberger, Amaral, Glass, Hausdorff, Ivanov,
  Mark, Mietus, Moody, Peng, and Stanley}{Goldberger et~al\mbox{.}}{2000}]%
        {Goldberger2000PhysioBankPA}
\bibfield{author}{\bibinfo{person}{Ary~L. Goldberger}, \bibinfo{person}{Luis
  A.~Nunes Amaral}, \bibinfo{person}{L Glass}, \bibinfo{person}{Jeffrey~M.
  Hausdorff}, \bibinfo{person}{Plamen~Ch. Ivanov}, \bibinfo{person}{Roger~G.
  Mark}, \bibinfo{person}{Joseph~E. Mietus}, \bibinfo{person}{George~B. Moody},
  \bibinfo{person}{Chung-Kang Peng}, {and} \bibinfo{person}{Harry~Eugene
  Stanley}.} \bibinfo{year}{2000}\natexlab{}.
\newblock \showarticletitle{PhysioBank, PhysioToolkit, and PhysioNet:
  components of a new research resource for complex physiologic signals.}
\newblock \bibinfo{journal}{\emph{Circulation}}  \bibinfo{volume}{101 23}
  (\bibinfo{year}{2000}), \bibinfo{pages}{E215--20}.
\newblock


\bibitem[\protect\citeauthoryear{Greenwald}{Greenwald}{1986}]%
        {Greenwald1986TheDA}
\bibfield{author}{\bibinfo{person}{Scott~D Greenwald}.}
  \bibinfo{year}{1986}\natexlab{}.
\newblock \showarticletitle{The development and analysis of a ventricular
  fibrillation detector}.
\newblock


\bibitem[\protect\citeauthoryear{Hido, Id{\'e}, Kashima, Kubo, and
  Matsuzawa}{Hido et~al\mbox{.}}{2008}]%
        {Hido2008UnsupervisedCA}
\bibfield{author}{\bibinfo{person}{Shohei Hido}, \bibinfo{person}{Tsuyoshi
  Id{\'e}}, \bibinfo{person}{Hisashi Kashima}, \bibinfo{person}{Harunobu Kubo},
  {and} \bibinfo{person}{Hirofumi Matsuzawa}.} \bibinfo{year}{2008}\natexlab{}.
\newblock \showarticletitle{Unsupervised Change Analysis Using Supervised
  Learning}. In \bibinfo{booktitle}{\emph{Pacific-Asia Conference on Knowledge
  Discovery and Data Mining}}.
\newblock


\bibitem[\protect\citeauthoryear{Hwang, Kim, Kim, and Seah}{Hwang
  et~al\mbox{.}}{2010}]%
        {Hwang2010ASO}
\bibfield{author}{\bibinfo{person}{Inseok Hwang}, \bibinfo{person}{Sungwan
  Kim}, \bibinfo{person}{Youdan Kim}, {and} \bibinfo{person}{Chze~Eng Seah}.}
  \bibinfo{year}{2010}\natexlab{}.
\newblock \showarticletitle{A Survey of Fault Detection, Isolation, and
  Reconfiguration Methods}.
\newblock \bibinfo{journal}{\emph{IEEE Transactions on Control Systems
  Technology}}  \bibinfo{volume}{18} (\bibinfo{year}{2010}),
  \bibinfo{pages}{636--653}.
\newblock


\bibitem[\protect\citeauthoryear{Imani and Keogh}{Imani and Keogh}{2021}]%
        {ImaniMultiWindowFinderDA}
\bibfield{author}{\bibinfo{person}{Shima Imani} {and} \bibinfo{person}{Eamonn
  Keogh}.} \bibinfo{year}{2021}\natexlab{}.
\newblock \showarticletitle{Multi-Window-Finder: Domain Agnostic Window Size
  for Time Series Data}.
\newblock \bibinfo{journal}{\emph{MileTS'21: 7th KDD Workshop on Mining and
  Learning from Time Series}} (\bibinfo{year}{2021}).
\newblock


\bibitem[\protect\citeauthoryear{Karimov, Rabl, Katsifodimos, Samarev,
  Heiskanen, and Markl}{Karimov et~al\mbox{.}}{2019}]%
        {Karimov2019BenchmarkingDS}
\bibfield{author}{\bibinfo{person}{Jeyhun Karimov}, \bibinfo{person}{Tilmann
  Rabl}, \bibinfo{person}{Asterios Katsifodimos}, \bibinfo{person}{Roman~S.
  Samarev}, \bibinfo{person}{Henri Heiskanen}, {and} \bibinfo{person}{Volker
  Markl}.} \bibinfo{year}{2019}\natexlab{}.
\newblock \showarticletitle{Benchmarking Distributed Stream Data Processing
  Systems}.
\newblock \bibinfo{journal}{\emph{2018 IEEE 34th International Conference on
  Data Engineering (ICDE)}} (\bibinfo{year}{2019}),
  \bibinfo{pages}{1507--1518}.
\newblock


\bibitem[\protect\citeauthoryear{Kemp, Zwinderman, Tuk, Kamphuisen, and
  Oberye}{Kemp et~al\mbox{.}}{2000}]%
        {Kemp2000AnalysisOA}
\bibfield{author}{\bibinfo{person}{Bob Kemp}, \bibinfo{person}{Aeilko~H.
  Zwinderman}, \bibinfo{person}{Bert Tuk}, \bibinfo{person}{Hilbert A.~C.
  Kamphuisen}, {and} \bibinfo{person}{Josefien J.~L. Oberye}.}
  \bibinfo{year}{2000}\natexlab{}.
\newblock \showarticletitle{Analysis of a sleep-dependent neuronal feedback
  loop: the slow-wave microcontinuity of the EEG}.
\newblock \bibinfo{journal}{\emph{IEEE Transactions on Biomedical Engineering}}
   \bibinfo{volume}{47} (\bibinfo{year}{2000}), \bibinfo{pages}{1185--1194}.
\newblock


\bibitem[\protect\citeauthoryear{Keriven, Garreau, and Poli}{Keriven
  et~al\mbox{.}}{2018}]%
        {Keriven2018NEWMAAN}
\bibfield{author}{\bibinfo{person}{Nicolas Keriven}, \bibinfo{person}{Damien
  Garreau}, {and} \bibinfo{person}{Iacopo Poli}.}
  \bibinfo{year}{2018}\natexlab{}.
\newblock \showarticletitle{NEWMA: A New Method for Scalable Model-Free Online
  Change-Point Detection}.
\newblock \bibinfo{journal}{\emph{IEEE Transactions on Signal Processing}}
  \bibinfo{volume}{68} (\bibinfo{year}{2018}), \bibinfo{pages}{3515--3528}.
\newblock


\bibitem[\protect\citeauthoryear{Kifer, Ben-David, and Gehrke}{Kifer
  et~al\mbox{.}}{2004}]%
        {Kifer2004DetectingCI}
\bibfield{author}{\bibinfo{person}{Daniel Kifer}, \bibinfo{person}{Shai
  Ben-David}, {and} \bibinfo{person}{Johannes Gehrke}.}
  \bibinfo{year}{2004}\natexlab{}.
\newblock \showarticletitle{Detecting Change in Data Streams}. In
  \bibinfo{booktitle}{\emph{Very Large Data Bases Conference}}.
\newblock


\bibitem[\protect\citeauthoryear{Krishnamurthi, Kumar, Gopinathan, Nayyar, and
  Qureshi}{Krishnamurthi et~al\mbox{.}}{2020}]%
        {Krishnamurthi2020AnOO}
\bibfield{author}{\bibinfo{person}{Rajalakshmi Krishnamurthi},
  \bibinfo{person}{Adarsh Kumar}, \bibinfo{person}{Dhanalekshmi Gopinathan},
  \bibinfo{person}{Anand Nayyar}, {and} \bibinfo{person}{Basit Qureshi}.}
  \bibinfo{year}{2020}\natexlab{}.
\newblock \showarticletitle{An Overview of IoT Sensor Data Processing, Fusion,
  and Analysis Techniques}.
\newblock \bibinfo{journal}{\emph{Sensors (Basel, Switzerland)}}
  \bibinfo{volume}{20} (\bibinfo{year}{2020}).
\newblock


\bibitem[\protect\citeauthoryear{Laguna, Garcia-Olaya, and Borrajo}{Laguna
  et~al\mbox{.}}{2011}]%
        {Laguna2011ADS}
\bibfield{author}{\bibinfo{person}{Javier~Ortiz Laguna}, \bibinfo{person}{Angel
  Garcia-Olaya}, {and} \bibinfo{person}{Daniel Borrajo}.}
  \bibinfo{year}{2011}\natexlab{}.
\newblock \showarticletitle{A dynamic sliding window approach for activity
  recognition}. In \bibinfo{booktitle}{\emph{User Modeling, Adaptation, and
  Personalization}}.
\newblock


\bibitem[\protect\citeauthoryear{Lara and Labrador}{Lara and Labrador}{2013}]%
        {Lara2013ASO}
\bibfield{author}{\bibinfo{person}{Oscar~D. Lara} {and}
  \bibinfo{person}{Miguel~A. Labrador}.} \bibinfo{year}{2013}\natexlab{}.
\newblock \showarticletitle{A Survey on Human Activity Recognition using
  Wearable Sensors}.
\newblock \bibinfo{journal}{\emph{IEEE Communications Surveys \& Tutorials}}
  \bibinfo{volume}{15} (\bibinfo{year}{2013}), \bibinfo{pages}{1192--1209}.
\newblock


\bibitem[\protect\citeauthoryear{Levchenko, Kolev, Yagoubi, Akbarinia,
  Masseglia, Palpanas, Shasha, and Valduriez}{Levchenko et~al\mbox{.}}{2020}]%
        {Levchenko2020BestNeighborEE}
\bibfield{author}{\bibinfo{person}{Oleksandra Levchenko},
  \bibinfo{person}{Boyan Kolev}, \bibinfo{person}{Djamel~Edine Yagoubi},
  \bibinfo{person}{Reza Akbarinia}, \bibinfo{person}{Florent Masseglia},
  \bibinfo{person}{Themis Palpanas}, \bibinfo{person}{Dennis Shasha}, {and}
  \bibinfo{person}{Patrick Valduriez}.} \bibinfo{year}{2020}\natexlab{}.
\newblock \showarticletitle{BestNeighbor: efficient evaluation of kNN queries
  on large time series databases}.
\newblock \bibinfo{journal}{\emph{Knowledge and Information Systems}}
  \bibinfo{volume}{63} (\bibinfo{year}{2020}), \bibinfo{pages}{349 -- 378}.
\newblock


\bibitem[\protect\citeauthoryear{Liakos, Papakonstantinopoulou, and
  Kotidis}{Liakos et~al\mbox{.}}{2022}]%
        {Liakos2022ChimpEL}
\bibfield{author}{\bibinfo{person}{Panagiotis Liakos}, \bibinfo{person}{Katia
  Papakonstantinopoulou}, {and} \bibinfo{person}{Yannis Kotidis}.}
  \bibinfo{year}{2022}\natexlab{}.
\newblock \showarticletitle{Chimp: Efficient Lossless Floating Point
  Compression for Time Series Databases}.
\newblock \bibinfo{journal}{\emph{Proc. VLDB Endow.}}  \bibinfo{volume}{15}
  (\bibinfo{year}{2022}), \bibinfo{pages}{3058--3070}.
\newblock


\bibitem[\protect\citeauthoryear{Matsubara, Sakurai, and Faloutsos}{Matsubara
  et~al\mbox{.}}{2014}]%
        {Matsubara2014AutoPlaitAM}
\bibfield{author}{\bibinfo{person}{Yasuko Matsubara}, \bibinfo{person}{Yasushi
  Sakurai}, {and} \bibinfo{person}{Christos Faloutsos}.}
  \bibinfo{year}{2014}\natexlab{}.
\newblock \showarticletitle{AutoPlait: automatic mining of co-evolving time
  sequences}.
\newblock \bibinfo{journal}{\emph{Proceedings of the 2014 ACM SIGMOD
  International Conference on Management of Data}} (\bibinfo{year}{2014}).
\newblock


\bibitem[\protect\citeauthoryear{Moody and Mark}{Moody and Mark}{2001}]%
        {Moody2001TheIO}
\bibfield{author}{\bibinfo{person}{George~B. Moody} {and}
  \bibinfo{person}{Roger~G. Mark}.} \bibinfo{year}{2001}\natexlab{}.
\newblock \showarticletitle{The impact of the MIT-BIH Arrhythmia Database}.
\newblock \bibinfo{journal}{\emph{IEEE Engineering in Medicine and Biology
  Magazine}}  \bibinfo{volume}{20} (\bibinfo{year}{2001}),
  \bibinfo{pages}{45--50}.
\newblock


\bibitem[\protect\citeauthoryear{Mueen, Hamooni, and Estrada}{Mueen
  et~al\mbox{.}}{2014}]%
        {Mueen2014TimeSJ}
\bibfield{author}{\bibinfo{person}{Abdullah~Al Mueen}, \bibinfo{person}{Hossein
  Hamooni}, {and} \bibinfo{person}{Trilce Estrada}.}
  \bibinfo{year}{2014}\natexlab{}.
\newblock \showarticletitle{Time Series Join on Subsequence Correlation}.
\newblock \bibinfo{journal}{\emph{2014 IEEE International Conference on Data
  Mining}} (\bibinfo{year}{2014}), \bibinfo{pages}{450--459}.
\newblock


\bibitem[\protect\citeauthoryear{Munchmeyer, Bindi, Leser, and
  Tilmann}{Munchmeyer et~al\mbox{.}}{2020}]%
        {Munchmeyer2020TheTE}
\bibfield{author}{\bibinfo{person}{Jannes Munchmeyer}, \bibinfo{person}{Dino
  Bindi}, \bibinfo{person}{Ulf Leser}, {and} \bibinfo{person}{Frederik
  Tilmann}.} \bibinfo{year}{2020}\natexlab{}.
\newblock \showarticletitle{The transformer earthquake alerting model: a new
  versatile approach to earthquake early warning}.
\newblock \bibinfo{journal}{\emph{Geophysical Journal International}}
  (\bibinfo{year}{2020}).
\newblock


\bibitem[\protect\citeauthoryear{Nolle, Badura, Catlett, Bowser, and
  Sketch}{Nolle et~al\mbox{.}}{1986}]%
        {Nolle1986crei}
\bibfield{author}{\bibinfo{person}{FM Nolle}, \bibinfo{person}{FK Badura},
  \bibinfo{person}{JM Catlett}, \bibinfo{person}{RW Bowser}, {and}
  \bibinfo{person}{MH Sketch}.} \bibinfo{year}{1986}\natexlab{}.
\newblock \showarticletitle{CREI-GARD, a new concept in computerized arrhythmia
  monitoring systems}.
\newblock \bibinfo{journal}{\emph{Computers in Cardiology}}
  \bibinfo{volume}{13}, \bibinfo{number}{1} (\bibinfo{year}{1986}),
  \bibinfo{pages}{515--518}.
\newblock


\bibitem[\protect\citeauthoryear{Page}{Page}{1954}]%
        {Page1954CONTINUOUSIS}
\bibfield{author}{\bibinfo{person}{E.~S. Page}.}
  \bibinfo{year}{1954}\natexlab{}.
\newblock \showarticletitle{CONTINUOUS INSPECTION SCHEMES}.
\newblock \bibinfo{journal}{\emph{Biometrika}}  \bibinfo{volume}{41}
  (\bibinfo{year}{1954}), \bibinfo{pages}{100--115}.
\newblock


\bibitem[\protect\citeauthoryear{Pan and Tompkins}{Pan and Tompkins}{1985}]%
        {Pan1985ARQ}
\bibfield{author}{\bibinfo{person}{Jiapu Pan} {and} \bibinfo{person}{Willis~J.
  Tompkins}.} \bibinfo{year}{1985}\natexlab{}.
\newblock \showarticletitle{A Real-Time QRS Detection Algorithm}.
\newblock \bibinfo{journal}{\emph{IEEE Transactions on Biomedical Engineering}}
   \bibinfo{volume}{BME-32} (\bibinfo{year}{1985}), \bibinfo{pages}{230--236}.
\newblock


\bibitem[\protect\citeauthoryear{Pan, Wang, Wang, Wang, and Wang}{Pan
  et~al\mbox{.}}{2022}]%
        {Pan2022NLCSC}
\bibfield{author}{\bibinfo{person}{Shuye Pan}, \bibinfo{person}{Peng Wang},
  \bibinfo{person}{Chen Wang}, \bibinfo{person}{Wei Wang}, {and}
  \bibinfo{person}{Jianmin Wang}.} \bibinfo{year}{2022}\natexlab{}.
\newblock \showarticletitle{NLC: Search Correlated Window Pairs on Long Time
  Series}.
\newblock \bibinfo{journal}{\emph{Proc. VLDB Endow.}}  \bibinfo{volume}{15}
  (\bibinfo{year}{2022}), \bibinfo{pages}{1363--1375}.
\newblock


\bibitem[\protect\citeauthoryear{Paparrizos, Kang, Boniol, Tsay, Palpanas, and
  Franklin}{Paparrizos et~al\mbox{.}}{2022}]%
        {Paparrizos2022TSBUADAE}
\bibfield{author}{\bibinfo{person}{John Paparrizos}, \bibinfo{person}{Yuhao
  Kang}, \bibinfo{person}{Paul Boniol}, \bibinfo{person}{Ruey Tsay},
  \bibinfo{person}{Themis Palpanas}, {and} \bibinfo{person}{Michael~J.
  Franklin}.} \bibinfo{year}{2022}\natexlab{}.
\newblock \showarticletitle{TSB-UAD: An End-to-End Benchmark Suite for
  Univariate Time-Series Anomaly Detection}.
\newblock \bibinfo{journal}{\emph{Proc. VLDB Endow.}}  \bibinfo{volume}{15}
  (\bibinfo{year}{2022}), \bibinfo{pages}{1697--1711}.
\newblock


\bibitem[\protect\citeauthoryear{Pelkonen, Franklin, Cavallaro, Huang, Meza,
  Teller, and Veeraraghavan}{Pelkonen et~al\mbox{.}}{2015}]%
        {Pelkonen2015GorillaAF}
\bibfield{author}{\bibinfo{person}{Tuomas Pelkonen}, \bibinfo{person}{Scott
  Franklin}, \bibinfo{person}{Paul Cavallaro}, \bibinfo{person}{Qi Huang},
  \bibinfo{person}{Justin Meza}, \bibinfo{person}{Justin Teller}, {and}
  \bibinfo{person}{Kaushik Veeraraghavan}.} \bibinfo{year}{2015}\natexlab{}.
\newblock \showarticletitle{Gorilla: A Fast, Scalable, In-Memory Time Series
  Database}.
\newblock \bibinfo{journal}{\emph{Proc. VLDB Endow.}}  \bibinfo{volume}{8}
  (\bibinfo{year}{2015}), \bibinfo{pages}{1816--1827}.
\newblock


\bibitem[\protect\citeauthoryear{Rakthanmanon, Campana, Mueen, Batista,
  Westover, Zhu, Zakaria, and Keogh}{Rakthanmanon et~al\mbox{.}}{2012}]%
        {Rakthanmanon2012SearchingAM}
\bibfield{author}{\bibinfo{person}{Thanawin Rakthanmanon},
  \bibinfo{person}{Bilson J.~L. Campana}, \bibinfo{person}{Abdullah~Al Mueen},
  \bibinfo{person}{Gustavo E. A. P.~A. Batista}, \bibinfo{person}{M.~Brandon
  Westover}, \bibinfo{person}{Qiang Zhu}, \bibinfo{person}{Jesin Zakaria},
  {and} \bibinfo{person}{Eamonn~J. Keogh}.} \bibinfo{year}{2012}\natexlab{}.
\newblock \showarticletitle{Searching and Mining Trillions of Time Series
  Subsequences under Dynamic Time Warping}.
\newblock \bibinfo{journal}{\emph{KDD : proceedings. International Conference
  on Knowledge Discovery \& Data Mining}}  \bibinfo{volume}{2012}
  (\bibinfo{year}{2012}), \bibinfo{pages}{262 -- 270}.
\newblock


\bibitem[\protect\citeauthoryear{Reiss and Stricker}{Reiss and
  Stricker}{2011}]%
        {Reiss2011TowardsGA}
\bibfield{author}{\bibinfo{person}{Attila Reiss} {and} \bibinfo{person}{Didier
  Stricker}.} \bibinfo{year}{2011}\natexlab{}.
\newblock \showarticletitle{Towards global aerobic activity monitoring}. In
  \bibinfo{booktitle}{\emph{PETRA '11}}.
\newblock


\bibitem[\protect\citeauthoryear{Reiss and Stricker}{Reiss and
  Stricker}{2012}]%
        {Reiss2012CreatingAB}
\bibfield{author}{\bibinfo{person}{Attila Reiss} {and} \bibinfo{person}{Didier
  Stricker}.} \bibinfo{year}{2012}\natexlab{}.
\newblock \showarticletitle{Creating and benchmarking a new dataset for
  physical activity monitoring}. In \bibinfo{booktitle}{\emph{PETRA '12}}.
\newblock


\bibitem[\protect\citeauthoryear{Santos, Tilly, Chandramouli, and
  Goldstein}{Santos et~al\mbox{.}}{2013}]%
        {Santos2013DiAlDS}
\bibfield{author}{\bibinfo{person}{Ivo Santos}, \bibinfo{person}{Marcel Tilly},
  \bibinfo{person}{Badrish Chandramouli}, {and} \bibinfo{person}{Jonathan
  Goldstein}.} \bibinfo{year}{2013}\natexlab{}.
\newblock \showarticletitle{DiAl: Distributed Streaming Analytics Anywhere,
  Anytime}.
\newblock \bibinfo{journal}{\emph{Proc. VLDB Endow.}}  \bibinfo{volume}{6}
  (\bibinfo{year}{2013}), \bibinfo{pages}{1386--1389}.
\newblock


\bibitem[\protect\citeauthoryear{Sch{\"a}fer, Ermshaus, and Leser}{Sch{\"a}fer
  et~al\mbox{.}}{2021}]%
        {Schfer2021ClaSPT}
\bibfield{author}{\bibinfo{person}{Patrick Sch{\"a}fer}, \bibinfo{person}{Arik
  Ermshaus}, {and} \bibinfo{person}{Ulf Leser}.}
  \bibinfo{year}{2021}\natexlab{}.
\newblock \showarticletitle{ClaSP - Time Series Segmentation}.
\newblock \bibinfo{journal}{\emph{Proceedings of the 30th ACM International
  Conference on Information \& Knowledge Management}} (\bibinfo{year}{2021}).
\newblock


\bibitem[\protect\citeauthoryear{Sch{\"a}fer and Leser}{Sch{\"a}fer and
  Leser}{2022}]%
        {Schfer2022MotifletsS}
\bibfield{author}{\bibinfo{person}{Patrick Sch{\"a}fer} {and}
  \bibinfo{person}{Ulf Leser}.} \bibinfo{year}{2022}\natexlab{}.
\newblock \showarticletitle{Motiflets - Simple and Accurate Detection of Motifs
  in Time Series}.
\newblock \bibinfo{journal}{\emph{Proc. VLDB Endow.}}  \bibinfo{volume}{16}
  (\bibinfo{year}{2022}), \bibinfo{pages}{725--737}.
\newblock


\bibitem[\protect\citeauthoryear{Schall-Zimmerman, Senobari, Funning,
  Papalexakis, Oymak, Brisk, and Keogh}{Schall-Zimmerman et~al\mbox{.}}{2019}]%
        {SchallZimmerman2019MatrixPX}
\bibfield{author}{\bibinfo{person}{Zachary Schall-Zimmerman},
  \bibinfo{person}{Nader~Shakibay Senobari}, \bibinfo{person}{Gareth~J.
  Funning}, \bibinfo{person}{Evangelos~E. Papalexakis}, \bibinfo{person}{Samet
  Oymak}, \bibinfo{person}{Philip Brisk}, {and} \bibinfo{person}{Eamonn~J.
  Keogh}.} \bibinfo{year}{2019}\natexlab{}.
\newblock \showarticletitle{Matrix Profile XVIII: Time Series Mining in the
  Face of Fast Moving Streams using a Learned Approximate Matrix Profile}.
\newblock \bibinfo{journal}{\emph{2019 IEEE International Conference on Data
  Mining (ICDM)}} (\bibinfo{year}{2019}), \bibinfo{pages}{936--945}.
\newblock


\bibitem[\protect\citeauthoryear{Schmidt, Reiss, D{\"u}richen, Marberger, and
  Laerhoven}{Schmidt et~al\mbox{.}}{2018}]%
        {Schmidt2018IntroducingWA}
\bibfield{author}{\bibinfo{person}{Philip Schmidt}, \bibinfo{person}{Attila
  Reiss}, \bibinfo{person}{Robert D{\"u}richen}, \bibinfo{person}{Claus
  Marberger}, {and} \bibinfo{person}{Kristof~Van Laerhoven}.}
  \bibinfo{year}{2018}\natexlab{}.
\newblock \showarticletitle{Introducing WESAD, a Multimodal Dataset for
  Wearable Stress and Affect Detection}.
\newblock \bibinfo{journal}{\emph{Proceedings of the 20th ACM International
  Conference on Multimodal Interaction}} (\bibinfo{year}{2018}).
\newblock


\bibitem[\protect\citeauthoryear{Thiese, Ronna, and Ott}{Thiese
  et~al\mbox{.}}{2016}]%
        {Thiese2016PVI}
\bibfield{author}{\bibinfo{person}{Matthew~S. Thiese},
  \bibinfo{person}{Brenden~B Ronna}, {and} \bibinfo{person}{Ulrike Ott}.}
  \bibinfo{year}{2016}\natexlab{}.
\newblock \showarticletitle{P value interpretations and considerations.}
\newblock \bibinfo{journal}{\emph{Journal of thoracic disease}}
  \bibinfo{volume}{8 9} (\bibinfo{year}{2016}), \bibinfo{pages}{E928--E931}.
\newblock


\bibitem[\protect\citeauthoryear{Triboan, Chen, Chen, and Wang}{Triboan
  et~al\mbox{.}}{2017}]%
        {Triboan2017SemanticSO}
\bibfield{author}{\bibinfo{person}{Darpan Triboan},
  \bibinfo{person}{Liming~Luke Chen}, \bibinfo{person}{Feng Chen}, {and}
  \bibinfo{person}{Zumin Wang}.} \bibinfo{year}{2017}\natexlab{}.
\newblock \showarticletitle{Semantic segmentation of real-time sensor data
  stream for complex activity recognition}.
\newblock \bibinfo{journal}{\emph{Personal and Ubiquitous Computing}}
  \bibinfo{volume}{21} (\bibinfo{year}{2017}), \bibinfo{pages}{411--425}.
\newblock


\bibitem[\protect\citeauthoryear{Truong, Oudre, and Vayatis}{Truong
  et~al\mbox{.}}{2020}]%
        {Truong2019SelectiveRO}
\bibfield{author}{\bibinfo{person}{Charles Truong}, \bibinfo{person}{Laurent
  Oudre}, {and} \bibinfo{person}{Nicolas Vayatis}.}
  \bibinfo{year}{2020}\natexlab{}.
\newblock \showarticletitle{Selective review of offline change point detection
  methods}.
\newblock \bibinfo{journal}{\emph{Signal Processing}}  \bibinfo{volume}{167}
  (\bibinfo{year}{2020}), \bibinfo{pages}{107299}.
\newblock


\bibitem[\protect\citeauthoryear{van~den Burg and Williams}{van~den Burg and
  Williams}{2020}]%
        {van2020evaluation}
\bibfield{author}{\bibinfo{person}{Gerrit~JJ van~den Burg} {and}
  \bibinfo{person}{Christopher~KI Williams}.} \bibinfo{year}{2020}\natexlab{}.
\newblock \showarticletitle{An evaluation of change point detection
  algorithms}.
\newblock \bibinfo{journal}{\emph{arXiv preprint arXiv:2003.06222}}
  (\bibinfo{year}{2020}).
\newblock


\bibitem[\protect\citeauthoryear{van Rijn and Hutter}{van Rijn and
  Hutter}{2017}]%
        {Rijn2017HyperparameterIA}
\bibfield{author}{\bibinfo{person}{Jan~N. van Rijn} {and}
  \bibinfo{person}{Frank Hutter}.} \bibinfo{year}{2017}\natexlab{}.
\newblock \showarticletitle{Hyperparameter Importance Across Datasets}.
\newblock \bibinfo{journal}{\emph{Proceedings of the 24th ACM SIGKDD
  International Conference on Knowledge Discovery \& Data Mining}}
  (\bibinfo{year}{2017}).
\newblock


\bibitem[\protect\citeauthoryear{Verma, Kawamoto, Fadlullah, Nishiyama, and
  Kato}{Verma et~al\mbox{.}}{2017}]%
        {Verma2017ASO}
\bibfield{author}{\bibinfo{person}{Shikhar Verma}, \bibinfo{person}{Yuichi
  Kawamoto}, \bibinfo{person}{Zubair~Md. Fadlullah}, \bibinfo{person}{Hiroki
  Nishiyama}, {and} \bibinfo{person}{Nei Kato}.}
  \bibinfo{year}{2017}\natexlab{}.
\newblock \showarticletitle{A Survey on Network Methodologies for Real-Time
  Analytics of Massive IoT Data and Open Research Issues}.
\newblock \bibinfo{journal}{\emph{IEEE Communications Surveys \& Tutorials}}
  \bibinfo{volume}{19} (\bibinfo{year}{2017}), \bibinfo{pages}{1457--1477}.
\newblock


\bibitem[\protect\citeauthoryear{Wan, O'Grady, and O'Hare}{Wan
  et~al\mbox{.}}{2015}]%
        {Wan2015DynamicSE}
\bibfield{author}{\bibinfo{person}{Jie Wan}, \bibinfo{person}{Michael~J.
  O'Grady}, {and} \bibinfo{person}{Gregory M.~P. O'Hare}.}
  \bibinfo{year}{2015}\natexlab{}.
\newblock \showarticletitle{Dynamic sensor event segmentation for real-time
  activity recognition in a smart home context}.
\newblock \bibinfo{journal}{\emph{Personal and Ubiquitous Computing}}
  \bibinfo{volume}{19} (\bibinfo{year}{2015}), \bibinfo{pages}{287--301}.
\newblock


\bibitem[\protect\citeauthoryear{Wen, Gao, Song, Sun, Xu, and Zhu}{Wen
  et~al\mbox{.}}{2019}]%
        {Wen_Gao_Song_Sun_Xu_Zhu_2019}
\bibfield{author}{\bibinfo{person}{Qingsong Wen}, \bibinfo{person}{Jingkun
  Gao}, \bibinfo{person}{Xiaomin Song}, \bibinfo{person}{Liang Sun},
  \bibinfo{person}{Huan Xu}, {and} \bibinfo{person}{Shenghuo Zhu}.}
  \bibinfo{year}{2019}\natexlab{}.
\newblock \showarticletitle{RobustSTL: A Robust Seasonal-Trend Decomposition
  Algorithm for Long Time Series}.
\newblock \bibinfo{journal}{\emph{Proceedings of the AAAI Conference on
  Artificial Intelligence}} \bibinfo{volume}{33}, \bibinfo{number}{01}
  (\bibinfo{date}{Jul.} \bibinfo{year}{2019}), \bibinfo{pages}{5409--5416}.
\newblock
\urldef\tempurl%
\url{https://doi.org/10.1609/aaai.v33i01.33015409}
\showDOI{\tempurl}


\bibitem[\protect\citeauthoryear{Wingerath, Gessert, Friedrich, and
  Ritter}{Wingerath et~al\mbox{.}}{2016}]%
        {Wingerath2016RealtimeSP}
\bibfield{author}{\bibinfo{person}{Wolfram Wingerath}, \bibinfo{person}{Felix
  Gessert}, \bibinfo{person}{Steffen Friedrich}, {and} \bibinfo{person}{Norbert
  Ritter}.} \bibinfo{year}{2016}\natexlab{}.
\newblock \showarticletitle{Real-time stream processing for Big Data}.
\newblock \bibinfo{journal}{\emph{it - Information Technology}}
  \bibinfo{volume}{58} (\bibinfo{year}{2016}), \bibinfo{pages}{186 -- 194}.
\newblock


\bibitem[\protect\citeauthoryear{Woollam, Munchmeyer, Tilmann, Rietbrock,
  Lange, Bornstein, Diehl, Giunchi, Haslinger, Jozinovi'c, Michelini, Saul, and
  Soto}{Woollam et~al\mbox{.}}{2022}]%
        {Woollam2022SeisBenchATF}
\bibfield{author}{\bibinfo{person}{J.~H. Woollam}, \bibinfo{person}{Jannes
  Munchmeyer}, \bibinfo{person}{Frederik Tilmann}, \bibinfo{person}{Andreas
  Rietbrock}, \bibinfo{person}{Dietrich Lange}, \bibinfo{person}{Thomas
  Bornstein}, \bibinfo{person}{Tobias Diehl}, \bibinfo{person}{Carlo Giunchi},
  \bibinfo{person}{Florian Haslinger}, \bibinfo{person}{Dario Jozinovi'c},
  \bibinfo{person}{Alberto Michelini}, \bibinfo{person}{Joachim Saul}, {and}
  \bibinfo{person}{Hugo Soto}.} \bibinfo{year}{2022}\natexlab{}.
\newblock \showarticletitle{SeisBench—A Toolbox for Machine Learning in
  Seismology}.
\newblock \bibinfo{journal}{\emph{Seismological Research Letters}}
  (\bibinfo{year}{2022}).
\newblock


\bibitem[\protect\citeauthoryear{Yamanishi and Takeuchi}{Yamanishi and
  Takeuchi}{2002}]%
        {Yamanishi2002AUF}
\bibfield{author}{\bibinfo{person}{Kenji Yamanishi} {and}
  \bibinfo{person}{Jun’ichi Takeuchi}.} \bibinfo{year}{2002}\natexlab{}.
\newblock \showarticletitle{A unifying framework for detecting outliers and
  change points from non-stationary time series data}.
\newblock \bibinfo{journal}{\emph{Proceedings of the eighth ACM SIGKDD
  international conference on Knowledge discovery and data mining}}
  (\bibinfo{year}{2002}).
\newblock


\bibitem[\protect\citeauthoryear{Yeh, Zhu, Ulanova, Begum, Ding, Dau, Silva,
  Mueen, and Keogh}{Yeh et~al\mbox{.}}{2016}]%
        {Yeh2016MatrixPI}
\bibfield{author}{\bibinfo{person}{Chin-Chia~Michael Yeh}, \bibinfo{person}{Yan
  Zhu}, \bibinfo{person}{Liudmila Ulanova}, \bibinfo{person}{Nurjahan Begum},
  \bibinfo{person}{Yifei Ding}, \bibinfo{person}{Hoang~Anh Dau},
  \bibinfo{person}{Diego~Furtado Silva}, \bibinfo{person}{Abdullah~Al Mueen},
  {and} \bibinfo{person}{Eamonn~J. Keogh}.} \bibinfo{year}{2016}\natexlab{}.
\newblock \showarticletitle{Matrix Profile I: All Pairs Similarity Joins for
  Time Series: A Unifying View That Includes Motifs, Discords and Shapelets}.
\newblock \bibinfo{journal}{\emph{2016 IEEE 16th International Conference on
  Data Mining (ICDM)}} (\bibinfo{year}{2016}), \bibinfo{pages}{1317--1322}.
\newblock


\bibitem[\protect\citeauthoryear{Zhang, Alghamdi, Zhang, Eltabakh, and
  Rundensteiner}{Zhang et~al\mbox{.}}{2022}]%
        {Zhang2022PARROTPC}
\bibfield{author}{\bibinfo{person}{Liang Zhang}, \bibinfo{person}{Noura~A.
  Alghamdi}, \bibinfo{person}{Huayi Zhang}, \bibinfo{person}{Mohamed~Y.
  Eltabakh}, {and} \bibinfo{person}{Elke~A. Rundensteiner}.}
  \bibinfo{year}{2022}\natexlab{}.
\newblock \showarticletitle{PARROT: pattern-based correlation exploitation in
  big partitioned data series}.
\newblock \bibinfo{journal}{\emph{The VLDB Journal}} (\bibinfo{year}{2022}).
\newblock


\bibitem[\protect\citeauthoryear{Zhu, Schall-Zimmerman, Senobari, Yeh, Funning,
  Mueen, Brisk, and Keogh}{Zhu et~al\mbox{.}}{2016}]%
        {Zhu2016MatrixPI}
\bibfield{author}{\bibinfo{person}{Yan Zhu}, \bibinfo{person}{Zachary
  Schall-Zimmerman}, \bibinfo{person}{Nader~Shakibay Senobari},
  \bibinfo{person}{Chin-Chia~Michael Yeh}, \bibinfo{person}{Gareth~J. Funning},
  \bibinfo{person}{Abdullah~Al Mueen}, \bibinfo{person}{Philip Brisk}, {and}
  \bibinfo{person}{Eamonn~J. Keogh}.} \bibinfo{year}{2016}\natexlab{}.
\newblock \showarticletitle{Matrix Profile II: Exploiting a Novel Algorithm and
  GPUs to Break the One Hundred Million Barrier for Time Series Motifs and
  Joins}.
\newblock \bibinfo{journal}{\emph{2016 IEEE 16th International Conference on
  Data Mining (ICDM)}} (\bibinfo{year}{2016}), \bibinfo{pages}{739--748}.
\newblock


\bibitem[\protect\citeauthoryear{Zhu, Schall-Zimmerman, Senobari, Yeh, Funning,
  Mueen, Brisk, and Keogh}{Zhu et~al\mbox{.}}{2017}]%
        {Zhu2017ExploitingAN}
\bibfield{author}{\bibinfo{person}{Yan Zhu}, \bibinfo{person}{Zachary
  Schall-Zimmerman}, \bibinfo{person}{Nader~Shakibay Senobari},
  \bibinfo{person}{Chin-Chia~Michael Yeh}, \bibinfo{person}{Gareth~J. Funning},
  \bibinfo{person}{Abdullah~Al Mueen}, \bibinfo{person}{Philip Brisk}, {and}
  \bibinfo{person}{Eamonn~J. Keogh}.} \bibinfo{year}{2017}\natexlab{}.
\newblock \showarticletitle{Exploiting a novel algorithm and GPUs to break the
  ten quadrillion pairwise comparisons barrier for time series motifs and
  joins}.
\newblock \bibinfo{journal}{\emph{Knowledge and Information Systems}}
  \bibinfo{volume}{54} (\bibinfo{year}{2017}), \bibinfo{pages}{203--236}.
\newblock


\end{thebibliography}

\end{document}